\documentclass{article}

\usepackage[preprint,nonatbib]{neurips_data_2023}

\usepackage[utf8]{inputenc} 
\usepackage[T1]{fontenc}    
\usepackage{hyperref}       
\usepackage{url}            
\usepackage{booktabs}       
\usepackage{amsfonts}       
\usepackage{nicefrac}       
\usepackage{microtype}      
\usepackage{xcolor}         

\usepackage{booktabs}

\usepackage{times}

\usepackage{amsmath}
\usepackage[ruled,linesnumbered]{algorithm2e}
\usepackage[noend]{algpseudocode}

\usepackage{microtype}
\usepackage{graphicx} 
\usepackage{bmpsize}
\usepackage{subfigure}
\usepackage{booktabs} 
\usepackage{float}

\usepackage{amsthm}
\usepackage{amssymb}
\usepackage{makecell}
\usepackage{bbm}

\usepackage{times}
\usepackage{caption}

\usepackage{ulem}
\usepackage{mathtools}

\newcommand{\indic}{\mathbbm{1}}

\newcommand{\expect}[2]{\mathbb{E}_{#1}\left[ #2 \right]}
\newcommand{\ex}[1]{\mathbb{E}\left[ #1 \right]}

\newcommand{\set}[1]{\left\{#1\right\}}

\newcommand{\env}{AdCraft}

\newcommand{\seqi}[3]{#1_{#3,0},\,#1_{#3,1}\,\dots,\,#1_{#3,#2}}

\let\originalleft\left
\let\originalright\right
\renewcommand{\left}{\mathopen{}\mathclose\bgroup\originalleft}
\renewcommand{\right}{\aftergroup\egroup\originalright}

\DeclarePairedDelimiter\paren{(}{)}

\usepackage[textsize=footnotesize,color=green!40]{todonotes}

\title{\env{}: An Advanced Reinforcement Learning Benchmark Environment for Search Engine Marketing Optimization}

\author{%
  Maziar Gomrokchi \thanks{Equal contribution}~~\thanks{Correspondence to: mgomrokchi@gmail.com}~, Owen Levin $^*$, Jeffrey Roach $^*$, Jonah White\\ 
  Automation and Optimization Group\\ System1\\
  \texttt{maziar.gomrokchi,owen.levin,jeff,jonah.white@system1.com} \\
}

\begin{document}

\maketitle

\begin{abstract}

We introduce \env{}, a novel benchmark environment for the Reinforcement Learning (RL) community distinguished by its stochastic and non-stationary properties. The environment simulates bidding and budgeting dynamics within Search Engine Marketing (SEM), a digital marketing technique utilizing paid advertising to enhance the visibility of websites on search engine results pages (SERPs). The performance of SEM advertisement campaigns depends on several factors, including keyword selection, ad design, bid management, budget adjustments, and performance monitoring. Deep RL recently emerged as a potential strategy to optimize campaign profitability within the complex and dynamic landscape of SEM, but it requires substantial data, which may be costly or infeasible to acquire in practice. Our customizable environment enables practitioners to assess and enhance the robustness of RL algorithms pertinent to SEM bid and budget management without such costs. Through a series of experiments within the environment, we demonstrate the challenges imposed by sparsity and non-stationarity on agent convergence and performance. We hope these challenges further encourage discourse and development around effective strategies for managing real-world uncertainties.

\end{abstract}

\section{Introduction} \label{sec:intro}
Search Engine Marketing (SEM) is a digital marketing approach designed to enhance the visibility of web pages on Search Engine Results Pages (SERPs) through paid advertising. 
SEM focuses advertising efforts on prospective customers actively seeking products or services on search engines. 
Effective advertising requires keyword research, bid management, budget management, and performance tracking to ensure ads reach the most relevant audience and effectively drive traffic and conversions. Keyword research entails identifying the most pertinent and popular search terms used by the target audience.
Bid and budget management in SEM encompasses the process of determining suitable bids for selected keywords, ensuring that ads are displayed to the appropriate audience, at the optimal time, and for the right price. 
Performance tracking involves monitoring ad performance through key metrics such as click-through rates (CTR), conversion rates, and other relevant indicators, subsequently optimizing campaigns based on these insights. Effective SEM campaigns entail a continuous cycle of testing, optimization, and refinement to confirm that ads accomplish business objectives such as scaling website traffic according to profitability.

In recent years, Deep RL has been fruitfully employed in a broad spectrum of fields, including but not limited to, digital marketing and advertising~\cite{cai2017real, zhao2019deep, zhao2018deep, Jeunen2022_AuctionGym, jin2018real, wu2018budget}. 
Specifically, Deep RL has shown substantial promise in bolstering the effectiveness of SEM campaigns through several distinct avenues. 
First, in terms of bidding strategies, SEM significantly hinges on bidding for keywords with the objective of placing ads in front of users conducting those specific search queries. 
Here, Deep RL algorithms can optimally tailor the bidding strategy, aligning it with the unique goals of the campaign, such as the maximization of clicks or conversions. Second, as SEM campaigns are frequently bounded by limited budgets, judicious budget allocation becomes paramount. 
In this context, Deep RL algorithms can play a vital role in determining the optimal budget distribution across disparate campaigns and keywords, thus enhancing the return on investment (ROI). 
Third, Deep RL algorithms have the potential to optimize SEM campaigns in real-time, facilitating dynamic adjustments to bids and budget allocations based on the live performance metrics of the campaign.

Deep RL in SEM demands ample training data and iterative processes for efficient performance, with unique challenges stemming from fluctuating traffic volumes, competitive auction costs, and seasonal revenue shifts. 
Solutions include continuous model weight adaptations, multiple algorithms for varying traffic types, and rigorous live production training. 
However, the daily aggregation of cost and revenue data tends to obscure detailed auction dynamics, adding complexity to the fine-tuning process. 
In the non-stationary environments common in fields like financial markets, online advertising, and robotics, RL agents grapple with optimizing reward signals amid changing circumstances. Environmental dynamics like wear and tear, shifting lighting, or other factors can cause optimal policies to fluctuate, complicating the learning of stable policies. Contrary to the physics-based environments like MuJuCo \cite{todorov2012mujoco}, which are somewhat simplified and fail to adequately represent the intricate dynamics of real-world environments, or the game of 2048 \cite{antonoglou2022planning}, which, while providing an engaging logic game, lacks practical application from a solution standpoint, SEM environments manifest pronounced stochasticity and non-stationarity. These attributes make them a more relevant and challenging testbed for developing and testing RL algorithms. We summarize our main contributions as follows:
\begin{enumerate}
    \item We introduce \env{}\footnote{\footnotesize Code available at \url{https://github.com/Mikata-Project/adcraft}.}, a family of parameterizable environments simulating SEM bidding and budgeting. This environment provides a realistic and complex setting that accurately represents challenges faced in real-world SEM applications.
    \item Our proposed environment encapsulates the non-stationary and stochastic nature of SEM, enabling RL researchers to explore algorithms that can adapt to such complexities. 
    \item The environment allows users to parameterize a wide variety of features to carefully tailor the stochastic outcomes and bidding scenarios to fit experimental needs.
    \item We conduct a rigorous evaluation of our SEM environment, including built-in metrics, demonstrating its efficacy in training RL agents for large-scale applications. The resultant benchmarks, inclusive of built-in baseline, provide a reference for refining and developing RL algorithms.
\end{enumerate}

\env{} mitigates the costs and time constraints of live production environments, fostering more feasible and impactful Deep RL research for dynamic real-world scenarios like SEM. It offers a realistic simulation that intrinsically incorporates non-stationarity, stochasticity, and various observation sparsity forms. We present \env{}, analyze its key features, and demonstrate their effects. We then compare experimental evaluations of state-of-the-art RL models on different \env{} parameterized regimes, highlighting the existing challenges these models face in such settings. Developing and utilizing realistic simulation environments like \env{} broadens the success of Deep RL algorithms in industrial settings, narrowing the gap between theoretical advances and practical applications.
\section{Related Work}\label{sec:related}

RL research has experienced significant growth in recent years \cite{mnih2015human, lillicrap2015continuous, silver2016mastering, franccois2018introduction, fawzi2022discovering}, driven in part by the development of various simulation environments \cite{mnih2013playing, tassa2018deepmind, brockman2016openai, wang2021alchemy}. These environments enable researchers to test and refine RL algorithms across diverse domains, including robotics, gaming, and SEM. In this section, we review the most notable environments designed for RL research in SEM, highlighting their key features and contributions. Cai et al. \cite{cui2011bid} present a simulation environment for online ad exchange marketplaces with a focus on bid landscape forecasting, a framework that combines Deep learning and feature engineering to model a dynamic bidding environment. Perlich et al. \cite{perlich2012bid} introduce an environment for targeted online advertising, emphasizing bid optimization and inventory scoring. They propose a hierarchical Bayesian model to estimate the probability distribution of user responses, allowing advertisers to optimize their bids. Zhang et al. \cite{zhang2014optimal} develop a simulation environment for display advertising, focusing on optimal real-time bidding. They propose a hierarchical bidding framework that models user response stochasticity and optimizes bid prices accordingly. Zhao et al.~\cite{zhao2018deep} create a real-time bidding environment for online advertising that simulates bidding on sponsored search keywords in real-time for each impression, despite aggregated data observations. The study employs Deep RL techniques to adapt to the inherent uncertainties in auction outcomes and user behavior. Unfortunately, the environment is not publicly accessible and requires multiple servers to run.

The publicly available environment most similar to our setting is AuctionGym~\cite{Jeunen2022_AuctionGym}. In AuctionGym, multiple agents each have a catalog of advertisements and are each presented with a partial observation of a random context vector sampled from a given distribution. Each agent must learn which ad to show a given context as well as how much to bid for an impression on the ad-context pair. Each agent has a private, known conversion value for each ad in their catalog. Every agent directly participates in every auction, and the results are logged by every agent. AuctionGym allows one to test the effect of various bid-learning strategies on both the auctioneer as well as the bidding agents in a competitive multi-agent context.

In SEM, bids are usually placed on intermediary platforms where campaigns targeting specific keywords are created, each determining a set of contexts for bidding. 
Accurately modeling the number of auctions per keyword requires more than just a probability distribution of contexts.
Platforms may differ in payment requirements, with some charging upon auction win and others only upon ad clicks.
They also often provide only aggregated data rather than individual auction outcomes. 
Furthermore, budget constraints, which influence auction outcomes and can limit an advertiser's ability to win auctions, are essential features in SEM. 
The constraints of AuctionGym \cite{Jeunen2022_AuctionGym} for modeling SEM bidding necessitate more than minor adjustments such as incorporating budgets or adding learnable parameters, as these only address a subset of existing discrepancies. 
With \env{}, we present a comprehensive SEM environment that directly addresses these discrepancies while allowing easy matching of simulation data directly to observed distributions in ongoing campaigns or explicit testing conditions to ensure robustness.
This approach promotes the exploration and advancement of RL methodologies within the complex, dynamic landscape of SEM.
\section{The \env{} Environment} \label{sec:sem-env}
\begin{figure*}
\centering
\includegraphics[width=0.9\textwidth]{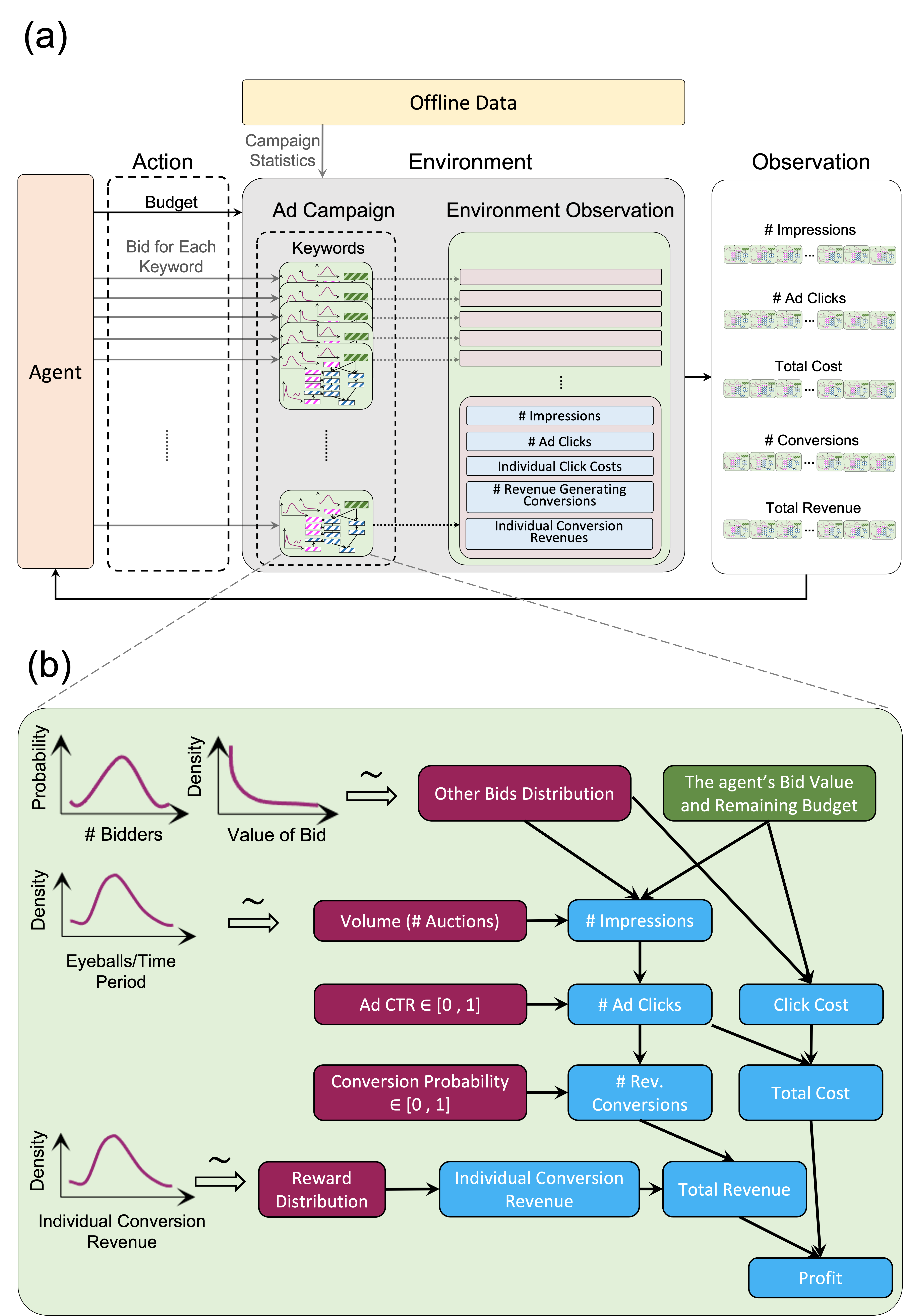} 
\caption{Schematic representation of the \env{} environment and the integration of keywords:
a) Illustration of an individual ImplicitKeyword's internal components interacting with an agent's bids. Source nodes (magenta) represent keyword-specific parameters that are concealed from the bidding agent. The environment processes non-source nodes (blue) and relays them to the agent.
b) The diagram demonstrates the agent submitting bids and budgets to the environment's collection of keywords. The budget is distributed across all keywords in the environment. The bidding and ad conversion outcomes are processed into observations that are returned to the agent.
}\label{fig:sem-arch}
\end{figure*}

\env{} offers the RL community a benchmarking platform for RL models in an SEM context. This section details the simulation environment, emphasizing its utility for efficient offline training in Deep RL for SEM tasks. Figure \ref{fig:sem-arch} presents the \env{} schematic.

\subsection{Task Overview}

In the \env{} environment, an agent manages an advertising campaign, distributing a shared budget across all campaign ads. The agent's main goal is to maximize a selected campaign metric, like profit. At every discrete time step, the agent decides the bid amount for each keyword. Each keyword samples some number of auctions. The agent's bid is submitted to each auction. When the agent's bid wins, it triggers an impression on the keyword and the corresponding ad displays. The agent pays the cost of winning a keyword auction when a user clicks on the ad. Still, we can simulate a pay-per-impression scenario by setting the click-through rate to one.

When a user clicks on an ad, a \textit{conversion} might follow, which brings revenue for the agent. Each keyword carries inherent probabilities of clicks and conversions and distributions for sampling the number of auctions and revenue amounts, among other parameters. If the agent exhausts its budget, it can no longer win auctions in that time step. Time steps in this context don't correspond to individual auctions. Instead, the agent receives observations aggregated over each time step, which might represent a day, hours, or minutes. With this aggregation in mind, we might write day and time steps interchangeably. The \env{} environment allows the agent to submit actions only after receiving each observation, structuring the decision-making process across discrete time intervals.

\subsection{Observations, actions, and task logic}
The environment is initialized with $K$ keywords, indexed by $k \in \set{0, 1, \dots, K-1}$, and a total number of days or time steps, $T$. At each time $t \in \set{0, 1, \dots, T-1}$, the agent submits a budget $B_t$ for that day's bidding, along with the keyword bids $b_t = (\seqi{b}{K-1}{t}) \in \mathbb{R}^K_{\geq 0.01}$ as an action to the environment. Upon submitting the action, the agent receives observations comprising the number of \textit{impressions} (auctions won) on each keyword $i_t \in \mathbb{Z}^K_{\geq 0}$, the number of \textit{ad} clicks on each keyword (times the agent paid the auction price) $a_t \in \mathbb{Z}^K_{\geq 0}$, the total money \textit{spent} on each keyword during the trial $s_t \in \mathbb{R}^K_{\geq 0}$, the number of \textit{revenue}-generating conversions on each keyword $r_t \in \mathbb{Z}^K_{\geq 0}$, and the total revenue \textit{earned} on each keyword $e_t \in \mathbb{R}^K_{\geq 0.01}$. These values are computed on a per-auction basis using the parameters described in Section~\ref{sec:keyword-flow} and aggregated to form the final per-keyword outputs. The agent also observes the cumulative profit of trials $\leq t$ and the trial number $t$ itself.

The environment provides an additional reward signal, which consists of the individual trial's net profit, along with two flags indicating whether the epoch of trials has concluded. Trials begin at 0 and increment until trial $T$ when the termination flag is set to True. Alternatively, trials may end early if the cumulative profit from trials falls below a threshold defined in the environmental parameters. This threshold can act as a meta-budget, which could end a campaign early if it consistently underperforms. The truncated flag is set to True when the cumulative profit falls below the threshold.

\subsubsection{Keyword flow}\label{sec:keyword-flow}  
Now, we describe the process of determining impressions, click costs, and revenue generation for each of the $K$ keywords in the \env{} simulation environment. For each keyword, $k$, the ``volume'', or the number of auctions, per trial is sampled from a distribution, and the bid $b_{t,k}$ is used in each of those auctions. The volume of auctions on a given keyword is sampled at the start of each time step from that keyword's internal volume distribution.

Impressions and click costs for each auction are determined using one of two methods depending on the keyword class: \textit{ImplicitKeyword} or \textit{ExplicitKeyword}. For ImplicitKeywords, the user bid $b_{t,k}$ participates in a literal auction against other bids sampled from an internal distribution. The auctions implicitly define a stochastic map from bid to cost and the number of impressions. In contrast, the ExplicitKeywords approach employs explicit, potentially non-deterministic functions that directly map $b_{t,k}$ to the probability of an impression and cost-per-click. After this step, both types of keywords follow an identical flow, with the probability of an ad click ($\Pr\paren{\text{ad click}}$) and the probability of conversion ($\Pr\paren{\text{conversion}\,\vert\,\text{ad click}}$) given an ad click being keyword-specific parameters that are not visible to the bidder. Upon each conversion, the revenue earned is sampled from a keyword-specific revenue distribution. To accelerate the computation of outcomes for every auction on each keyword, days are divided into $24$ sub-time steps. Within each sub-time step, outcomes for each keyword are vectorized and subsequently recombined. The remaining budget is recalculated after each set of sub-outcomes is recombined, and the trial's auctions may be terminated early if the budget is exhausted. A detailed example of how outcomes are computed when bidding on two keywords is provided in Appendix~\ref{app:bidding_example}.

\subsubsection{Augmenting with real-world data}
\env{} capitalizes on offline data to create realistic advertising scenarios.
Each parameter and distribution described in~\ref{sec:keyword-flow} can be either initialized by passing a list of keyword parameters or randomly sampled from chosen distributions. 
This allows for a fully customizable ad campaign to simulate arbitrary keyword popularity, competition, revenue sparsity, and more. 
We find this method of environment initialization affords a high measure of control over the simulation scenarios.

In one application, keywords are sampled to align with empirical observations through a joint parameter distribution fitting. This method preserves inter-parameter relationships, ensuring a simulation congruent with real ad campaign outcomes. We offer a tool for keyword initialization from specified product parameter quantiles, but intricate parameter couplings are not inherently supported.

\subsection{The environment non-stationarity}\label{sec:nonstationarity}

The \env{} environment seeks to emulate the dynamic and uncertain nature of live auction scenarios. To achieve this, it employs a non-stationary model where keyword-specific parameters fluctuate over time. Time-varying parameters can include the average number of auctions, click-through rates, and conversion rates for each keyword.  The environment allows users to tailor these functions to yield either a static value or a value drawn from a predefined distribution. By default, values for each keyword are drawn from keyword-specific distributions, which users can modify to suit their requirements.

Furthermore, the \env{} environment also features daily variations in certain parameters, thereby introducing a layer of non-stationarity. Parameters such as click-through rates, conversion rates, and average auction volumes can exhibit daily fluctuations. By default, \env{} utilizes a multiplicative random walk for click-through rate and conversion rate, where the respective value is multiplied by a random step between $1-\eta$ and $1+\eta$ for small positive $\eta$s selected by the user. The resultant values are clipped within $[0,1]$ to ensure valid probabilities. In contrast, changes in a keyword's mean volume are modelled as additive random walks with step sizes sampled uniformly from $[-\eta\cdot \mathrm{Vol}_{\mathrm{init}},\,\eta\cdot \mathrm{Vol}_\mathrm{init}]$. The final mean volume is clipped to remain non-negative. 
Figure~\ref{fig:four_figures}~(a) shows the random walks' effects on maximum expected profits. These features of non-stationarity and stochasticity imbue the \env{} environment with a dynamic representation of live auction scenarios, thereby enhancing the effectiveness of the simulation for testing and optimizing advertising strategies.
\section{Evaluation Methodology} \label{sec:exp}
In this section, we discuss the selection of a built-in baseline algorithm for comparison, along with three well-established Deep RL algorithms and two tailored evaluation metrics specifically designed for the \env{} environment. 

\subsection{Algorithms}
In our experimental approach, we first evaluate a built-in baseline algorithm provided by the \env{} environment, which serves as a crucial reference point for comparison. 
The baseline algorithm gradually increases bids on each keyword until outcomes are observed. After observing outcomes, the bidder bids proportionally to their estimated value of a single auction (see Appendix for further details). Vickrey (1952) demonstrated in \cite{vickrey_second_price} that this strategy would be optimal in our setting if two assumptions are met: (1) bids do not change the value of the ad, and (2) the value of the ad is already known to the agent. However, both of these assumptions are violated. Firstly, the total revenue is proportional to ad clicks, which increases with the bid, but the relationship between bid and ad clicks may not be linear. Secondly, initially, the model lacks revenue or conversion rate data for a given advertisement, resulting in uncertainty regarding the value of the ad. With an accurate-enough estimate of ad value, the hurdle of (2) is mollified. In cases where only assumption (1) is violated, bidding the true value of a single auction guarantees profitability, although it does not optimize the consideration of impressions and clicks that different bids might generate. Given that the value of ads for each bid on each keyword must be learned during the bidding process, an important challenge lies in efficiently acquiring optimal bids for advertisements on each keyword. While the baseline model relies on a simple heuristic, our objective is to train a series of Deep RL algorithms that can more efficiently learn bid and advertisement values. 

Subsequently, we scrutinized the performance of Proximal Policy Optimization (PPO) \cite{schulman2017proximal}, Twin Delayed Deep Deterministic Policy Gradient (TD3) \cite{fujimoto2018addressing}, and Advantage Actor-Critic (A2C) \cite{mnih2016asynchronous} algorithms within the \env{} environment. These algorithms represent a broad range of state-of-the-art Deep RL methods. PPO excels in handling continuous action spaces and provides stable policy updates. A2C combines actor-critic methods with asynchronous updates, making it suitable for balancing exploration and exploitation. TD3 is designed for high-dimensional continuous control problems, utilizing twin critics and delayed updates for improved stability. These algorithm choices allow us to showcase different features of the \env{} environment and assess the impact of sparsity and non-stationarity on Deep RL algorithm performance in SEM tasks.

\subsection{Evaluation Metrics}

In practice many different performance indicators are used in SEM bidding, such as \textit{total clicks}, \textit{return on ad spend}, \textit{net profit}, and more. We focus our investigations on cumulative net profit of bidding over a $60$ time step window representing a two month advertisement campaign.
However, different ad campaigns might vary wildly in terms of profitability causing difficulty comparing campaign performance using only profit.  To that end, the key metrics we focus on normalize profits by taking the ratio of an agents profit with the expected profit of optimal bidding. This normalized cumulative profit (NCP) can be determined either wholistically looking at the total profit of a campaign, or on a per-keyword basis, where we average per keywords the NCP achieved on each. Average per-Keyword NCP (AKNCP) will be the second metric we examine in our experiments. For precise definitions of these metrics, refer to equation \eqref{eq:NCP} and \eqref{eq:AKNCP} in the Appendix respectively.

For each metric, the score is bounded above in expectation by $1$, but `lucky' bidding could push them higher. Neither metric is explicitly bounded below if a model experiences loss on average per keyword or overall during a campaign, but both are finite due to the finite number of auctions and finite bid sizes.
Typical values for an effective bidding strategy will lie between $0$ and $1$ for both metrics, with an NCP near $1$ indicating the the model had near optimal profit. A high AKNCP value signifies effective optimization across a majority of keywords, while a high NCP accompanied by a low AKNCP might indicate missed opportunities on certain keywords, but primarily the less profitable ones.

\section{Experiments and Discussion} \label{sec:discussion}

This section presents a range of experimental scenarios we developed to assess the effects of various parameter choices within \env{} using our baseline algorithm. We then benchmark the performance of Deep RL algorithms within different \env{} parameter regimes. These setups highlight the inherent complexity, stochasticity, and non-stationarity of the \env{} environment. It is pertinent to mention that our experiments inherently possess stochasticity due to the role of keywords' internal distributions in shaping \textit{auction volume}, \textit{impressions}, \textit{clicks}, \textit{costs}, \textit{conversions}, and \textit{revenue}. Nevertheless, if preferred, these parameters can be initialized with deterministic functions, thereby eliminating stochasticity (see Appendix for further details). All three Deep RL agents were trained using $500$ training iterations for approximately $1e6$ time-steps, ensuring equal environmental exposure. Further details regarding Deep RL agents' model architectures are presented in separate tables in Appendix. 

\subsection{Impact of Sparsity on Model Performance}
Real-world keywords are often ``long-tail," e.g.,  receiving traffic only rarely or traffic generating revenue only rarely. Agents bidding competently in these sparse-reward scenarios is paramount to their success. Here we evaluate the effects of sparsity within \env{} using a baseline algorithm, and we benchmark Deep RL algorithms' performances under various sparsity regimes. The \env{} environment features three notions of sparsity: \textit{keyword volume}, \textit{click-through rate} (CTR), and \textit{conversion rate} (CVR). These factors hierarchically influence the sparsity of rewards in SEM bidding. Specifically, keyword volume determines the number of auctions conducted for a given keyword. Auction outcomes are only observed when ads win and receive impressions, resulting in a low volume leading to a limited number of observed outcomes for a specific keyword. CTR signifies the probability of an ad being clicked given an impression. Even if a substantial number of auctions are won, a low CTR yields few observed costs, and only a subset of those costs generates revenue. CVR represents the probability of earning revenue from a clicked ad. The reward, defined as profit, is computed as the difference between revenue and cost. Hence, the hierarchy of these features actively shapes the sparsity structure of the observed rewards landscape in the SEM environment.

\begin{figure}[htbp]
    \begin{center}
    \includegraphics[width=1.0\textwidth]{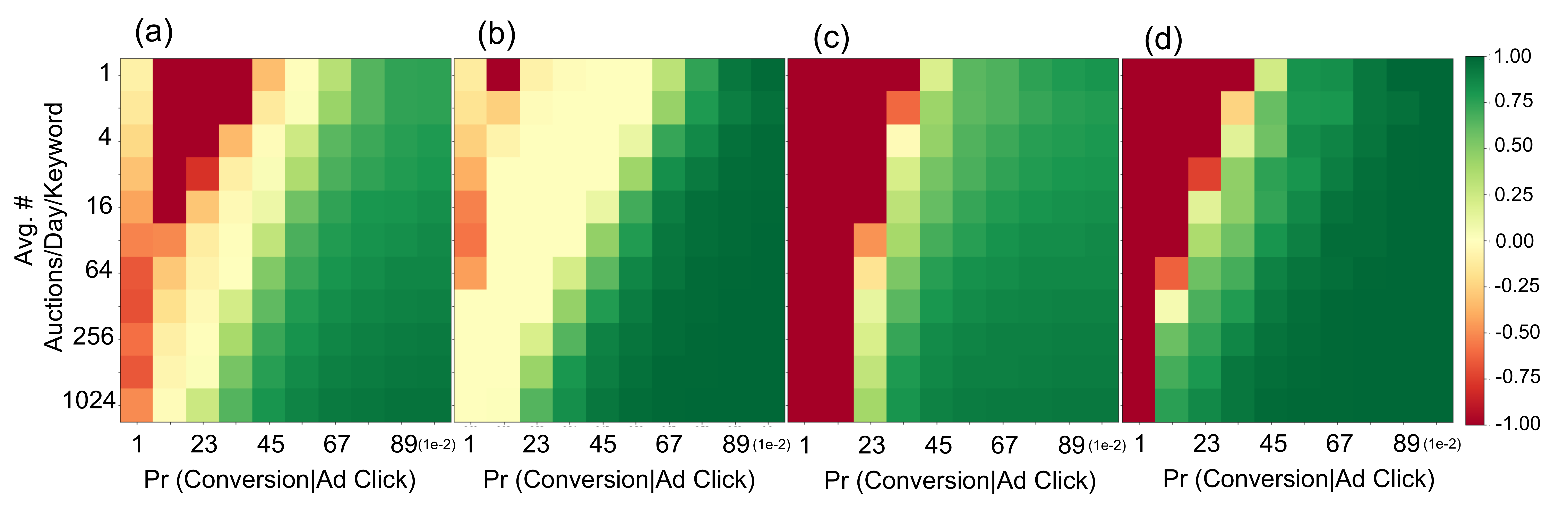} 
    \end{center}
    \caption{
        Signed heatmaps of AKNCP and NCP depict the performance of the baseline bidder across campaigns, with fixed pairs of auction volume and conversion rate. Scores below -1 are truncated to -1. The figure includes (a) median AKNCP from days 0 to 60, (b) median AKNCP from days 30 to 60, (c) median NCP from days 0 to 60, and (d) median NCP from days 30 to 60. Results represent the median score over 16 seeds. See the Appendix for maximum and minimum performances.
        }
        \label{fig:sparsity_heatmaps}
\end{figure}

To assess the impact of various choices of these features, we demonstrate the success and failure of the baseline model on different fixed choices of these features while keeping the remaining joint distribution of parameters otherwise identical. Each experiment runs on a 100-keyword sample from the parameter distributions. Figure~\ref{fig:sparsity_heatmaps} shows the results that illustrate the interplay between daily volume and CVR, while other pairs of sparsity parameters' results and a description of the full parameter distribution for all our experiments can be found in the Appendix.
Each cell of every heatmap corresponds to a specific sparsity setting. In each row, the mean daily volume for each keyword remains fixed, allowing us to observe the conditional effect of changing the conversion rate. Similarly, we can observe the conditional effect of changing keyword volumes within each column of the fixed conversion rate. The score of each cell in Figure~\ref{fig:sparsity_heatmaps} reflects the baseline model's ability to estimate revenue accurately. It is important to note that accurate estimation might incur costs or require an extended period, especially for sparse keywords. In Figure~\ref{fig:four_figures}~(a), we present a contrast between the observed and expected profits for a sparse keyword's bid that maximizes profit, highlighting the inherent difficulty. A phase transition exists between the top-left regimes of high sparsity and the bottom-right regimes of relative density. To assess the performance of Deep RL algorithms trained on sparse and dense regimes, we select two cells on opposite sides of the phase transition for closer examination. In the dense regime, we set the mean daily volume for each keyword to $128$ and a conversion rate of $0.8$. In contrast, we focus on Deep RL algorithms trained in a very sparse regime with a keyword mean volume of $16$ and a conversion rate of $0.1$.

Analyzing the behavior of the Deep RL algorithms AKNCP and NCP metrics provides significant insights. AKNCP, sensitive to how effectively an agent bids across the array of keywords, underscores the depth of an algorithm's strategy. NCP, on the other hand, offers a holistic perspective, highlighting the algorithm's overall profit optimization capabilities.

\begin{figure}[htbp]
\begin{center}
\includegraphics[width=1.0\textwidth]{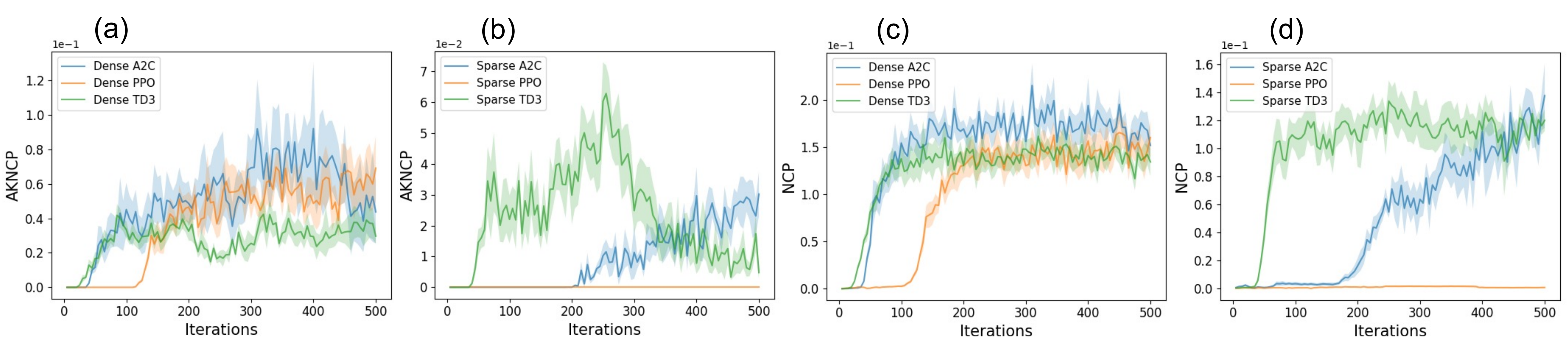} 
\end{center}
\caption{
Performance comparison of three Deep RL algorithms (A2C, PPO, and TD3) in dense and sparse settings with campaign size of $100$ keywords, showcasing their performance change. AKNCP and NCP performance metrics are utilized, with results averaged over five random seeds. Error bars represent the standard error on the mean. A2C and TD3 outperform PPO, indicating their superior handling of the exploration-exploitation tradeoff.}
\label{fig:RL_models_stationary}
\end{figure}

TD3's dominance in the NCP metric (\ref{fig:RL_models_stationary} (d)) resonates with its strategy to capitalize on high-reward pathways, occasionally at the expense of subtler profit opportunities. The deterministic policy gradient nature of TD3, coupled with its off-policy training, allows it to concentrate on distinct high-profit keyword trends. This focus, however, sometimes results in neglecting keywords that, while less profitable, could contribute to a more balanced performance, as reflected by the AKNCP metric (\ref{fig:RL_models_stationary} (b)). In juxtaposition, the on-policy nature of A2C equips it with versatility, enabling it to tap into a broader spectrum of keyword opportunities. This versatility becomes particularly salient in sparse environments (\ref{fig:RL_models_stationary} (b)), ensuring that A2C does not restrict its strategies to only high-value keywords but remains adaptable to the immediate feedback it receives.

In our setting, PPO's hybrid design, blending on-policy and off-policy elements, encounters challenges in high-dimensional, sparse environments \cite{pmlr-v80-kang18a}. The policy optimization nature of PPO further contributes to its under-performance in the sparse \env{} context. While its equilibrium aims for stable updates, it can hamper its performance, particularly when evaluated with AKNCP. This underscores the observed performance disparities between PPO, A2C, and TD3.

In summation, the intricate interplay of algorithmic characteristics with environmental sparsity and density, as illuminated by the AKNCP and NCP metrics, sheds light on the distinct strategies, strengths, and challenges faced by each Deep RL algorithm in our study. This analytical dive is especially pertinent in the context of \env{}. SEM, by its very nature, is a dynamic, high-dimensional, and oftentimes sparse arena, necessitating strategies that can adeptly balance between immediate gains and long-term profitability. As we navigate the multifaceted landscape of keyword bidding, campaign optimization, and budget allocation in SEM, the choice of algorithm becomes crucial. Our study on AdCraft underscores the imperative of aligning algorithmic capabilities with the idiosyncrasies of SEM. 

\subsection{Impact of Non-Stationarity on Model Performance}
\begin{figure}[htbp]
  \begin{center}
    \includegraphics[width=1.0\textwidth]{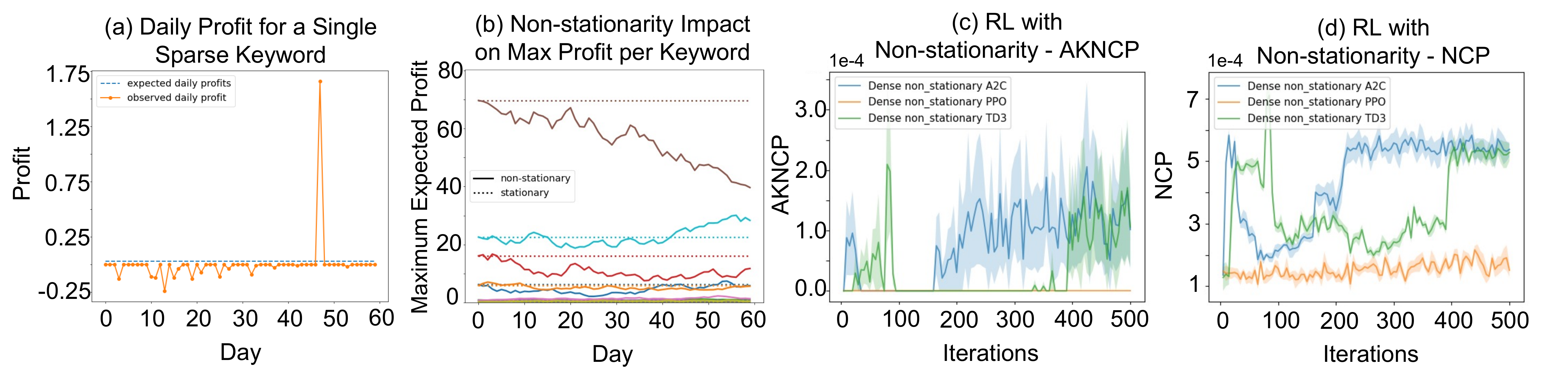}
    \end{center}
  \caption{(a) Profits for a stationary, sparse keyword's optimal bid. (b) Comparison of stationary and non-stationary optimal profits for ten dense keywords, demonstrating the impact of environmental dynamics on potential earnings. (c),(d) Performance assessment of three standard Deep RL algorithms - A2C, TD3, and PPO - in a dense, non-stationary regime with campaign size of $100$ keywords, focusing on AKNCP and NCP metrics, to highlight their adaptability under non-stationary environment dynamics. We average the results over $5$ random seeds and represent the error bars as the standard error on the mean.}
  \label{fig:four_figures}
\end{figure}

In this section, we conduct a series of experiments to examine the performance of the standard baseline and Deep RL algorithms in an environment with non-stationary transition dynamics. As we explain in Section~\ref{sec:nonstationarity}, non-stationary parameters in \env{} mimic the sparsity parameters. These parameters, such as \textit{mean volume}, \textit{CTR}, and \textit{CVR}, can change over time, making the SEM environment inherently non-stationary. Therefore, an algorithm's ability to adapt to these changes is critical to its usefulness in the SEM context. Figure~\ref{fig:four_figures}(b) demonstrates the impact of non-stationarity on the maximum expected profits for ten keywords from the dense regime. These keywords initially participate in an average of $128$ auctions per day with an initial CVR of $0.8$. Both values fluctuate over time, along with their CTR. This variability leads to notable shifts in outcomes within a single simulation run. When we move from a stationary to a non-stationary environment, the performance of Deep RL algorithms drops sharply, decreasing by three orders of magnitude. This is clear when comparing outcomes between Figures~\ref{fig:RL_models_stationary} and ~\ref{fig:four_figures}.  These results highlight the significant role of non-stationarity in the effectiveness of Deep RL solutions.

\section{Conclusion and Future Work} \label{sec:conclusion-future}
We introduce \env{}, an innovative simulation environment for SEM applications that bridges theoretical RL advancements and real-world applications. It incorporates key SEM attributes - non-stationarity, stochasticity, and complex real-world components. Our evaluation of various state-of-the-art Deep RL algorithms and the built-in baseline within \env{} reveals the extent of the challenges Deep RL faces when dealing with unique SEM problems such as non-iid data points, dynamic environments, online learning, and scalability. It also shines light on the current limitations of RL algorithms in adapting to environmental non-stationarity. The development of \env{} can be enhanced by including multiple agents, asynchronous actions and observations, non-uniform intraday auction volume, and accounting for ad placements' impact on bid outcomes. This platform opens the way for creating RL algorithms specifically designed for SEM. These advancements aim to improve SEM campaign optimization and management in practical applications.

\textbf{Limitations:} While \env{} aims to realistically simulate SEM dynamics, certain limitations remain. The environment models a simplified version of the competitive landscape, with stochastic bidder distributions rather than adaptive adversaries. Additionally, our experiments utilized a limited campaign size of $100$ keywords for computational tractability, whereas real-world campaigns may contain thousands. Factors like ad creatives, landing pages, and user journey impact conversion rates, which our current environment does not capture. We focused evaluation on short-term episodic returns, yet practical applications demand long-term continual learning. The simulated dynamics still only approximate real-world complexities. However, we believe \env{} provides a significant advance as a modular platform to incrementally integrate enhanced representations. Our experiments underscore challenges posed by non-stationarity and sparsity. But broader benchmarks on large-scale problems will likely reveal additional difficulties. Overall, while limitations exist, \env{} facilitates impactful research on sequential decision-making under uncertainty. We aim to incrementally enhance model representativeness and scope based on community feedback.


\section*{Acknowledgments and Disclosure of Funding}
The authors thank System1 LLC for funding and supporting this work.

\bibliography{References}
\bibliographystyle{abbrv}

\newpage

\appendix
\appendix 
\appendix
\appendix
\appendix
\appendix
\appendix
\appendix
\section{Repository and Documentation} \label{app:repo}

The repository for this project can be found at \url{https://github.com/Mikata-Project/adcraft}.

The documentation for this project can be found at \url{https://github.com/Mikata-Project/adcraft}.
\section{Reinforcement Learning and Search Engine Marketing Framework}\label{app:RL-SEM}

Deep Reinforcement Learning (RL) has earned recognition as a promising model for online advertising \cite{zhang2014optimal, cai2017real, zhao2018deep}, adept at navigating the unique challenges of this domain. Search Engine Marketing (SEM), with its rapidly changing environment, uncertain user behaviour, and seasonal effect, is a ripe area for RL applications. Algorithms such as Deep Q-Networks (DQN)~\cite{mnih2015human}, Soft Actor-Critic (SAC) \cite{haarnoja2018soft}, and Stochastic MuZero \cite{antonoglou2022planning} have demonstrated success in handling dynamic and uncertain environments, reinforcing their suitability for SEM.

The intricacies of SEM primarily entail sequential decision-making, with advertisers faced with time-sensitive decisions encompassing keyword selection, bid setting, and budget allocation. RL effectively models these issues as Markov Decision Processes (MDP) \cite{sutton2018reinforcement}, facilitating agile adaptation to environmental changes. A critical aspect of SEM involves balancing exploring new bidding strategies with exploiting profitable known ones. RL algorithms tackle this trade-off employing strategies such as $\epsilon$-greedy~\cite{sutton2018reinforcement}, upper confidence bound (UCB)\cite{auer2002finite}, and Thompson sampling\cite{thompson1933likelihood}. Further complexities in SEM arise due to the non-iid nature of data, with SEM data often displaying temporal and contextual correlations. RL techniques adeptly handle such dependencies, learning from continuous interactions. Scalability also becomes crucial as SEM complexity and potential actions increase. Deep RL techniques, including DQN\cite{mnih2015human}, Proximal Policy Optimization (PPO)\cite{schulman2017proximal}, and Actor-Critic methods~\cite{konda1999actor, haarnoja2018soft}, leverage deep learning for tackling large-scale problems. Their success across diverse SEM settings like real-time bidding~\cite{zhao2018deep}, bid optimization~\cite{zhang2014optimal}, and campaign management~\cite{cui2011bid} signal an increasing adoption of RL in SEM.

\section{Search Engine Marketing Dynamics and Challenges}\label{app:SEM_dynamics}

The SEM environment exhibits a dynamic nature characterized by several key factors. Firstly, the traffic volume for each unit fluctuates daily, with some units experiencing substantial variations and others exhibiting sparse observations. The competitive auction dynamics also contribute to the varying costs associated with each unit. Furthermore, revenue generation is influenced by significant seasonal effects, including daily budget resets, weekday versus weekend patterns, fluctuations at the end of the month, quarter, and year, and holidays. To make accurate business decisions, gather sufficient data on each traffic unit, which can range from as short as one day to as long as a year, depending on the combination of these factors. Given the system's constantly changing and noisy nature, continuous updating and adaptation of model weights are necessary, often requiring re-training or fine-tuning of algorithms. Moreover, the involvement of multiple algorithms to handle diverse traffic populations further amplifies the complexity of training and adaptation processes.

When placing a bid to advertise a specific traffic unit, the reporting of results may occur with a certain lag, preventing true real-time bidding unless one bids without full information. Each report provides aggregated data from potentially thousands of auctions, which limits the visibility into the detailed dynamics of the internal auction system. Setting bids too low can result in minimal ad visibility, while placing bids too high can lead to significantly increased costs exceeding the generated revenue. The aggregation of cost and revenue data exacerbates these challenges as the impact of a bid is only revealed after data is reported. In situations where bids are set well above profitable levels, substantial losses could occur without proper monitoring. Furthermore, disruptions, gaps, or delays in data can prolong the effects of unfavourable bids until data validity is restored. These complexities, combined with the feedback loops from various partners, contribute to the difficulty and cost associated with training RL algorithms in live production environments.

In the SEM environment, training RL algorithms is particularly challenging due to the dynamic and noisy nature of the system. The fluctuating traffic volume, varying costs, and seasonal effects introduce significant uncertainties that must be accounted for. Furthermore, the delayed reporting of results and the aggregated nature of the data further hinder the fine-grained analysis of the internal auction system. The bid optimization process requires careful consideration of trade-offs, avoiding bids that are too low or too high to maximize ad visibility and minimize costs. Additionally, disruptions in data and the influence of partner feedback loops add to the complexity of training RL algorithms effectively.

Overall, the SEM environment presents various dynamic challenges that impact the training and deployment of RL algorithms. Understanding and addressing these challenges are crucial for developing robust and effective solutions in search engine marketing.
\section{Implementation} \label{app:implementation}

\env{} is a simulation environment built upon the Farma Foundation Gymnasium (formerly known as OpenAI GYM), comprising a sequence of `trials' that collectively form `episodes'. Gymnasium, introduced by Brockman et al.\cite{brockman2016openai}, is a toolkit for the development and comparison of RL algorithms, offering an extensive array of environments, including classic control problems, Atari games, and robotic simulations. While not directly targeting SEM, Gymnasium has laid the groundwork for numerous RL research initiatives, spurring the creation of environments catering to specific use-cases, such as SEM. 

\env{} seamlessly integrates with the Gymnasium API, facilitating the inclusion of customized reinforcement learning libraries and algorithms. This adaptable framework empowers researchers to tailor their strategies based on specific requirements, preferences, and expertise, while leveraging the comprehensive capabilities of the environment.

The incorporation of the Rust programming language, renowned for its performance and memory efficiency, has substantially accelerated computationally intensive tasks such as sampling from distributions. Transitioning specific functions to Rust has resulted in performance enhancement of up to 110x speed for some functions. This contributed to an overall speedup of over 3x for each environment step. The significant improvements expedite training times, thereby enabling more efficient scaling in high-dimensional settings. However, this comes with the trade-off of increased complexity in environment packaging for seamless installation.
\section{Environment Parameters and Experiment Parameterization} \label{app:env_params}
Each keyword within the framework possesses internal probability distributions for sampling volume and revenue, along with floating-point values for click-through rate (CTR) and conversion rate (CVR).

During the initialization of a keyword set, there is flexibility in selecting the desired distributions. To facilitate the sampling of different distributions for each keyword, we use parametric distributions for these features in our experiments so that creating random distributions requires only sampling random parameters. The distributions from which we sample parameters are use-specified. In our experiments involving sets of ImplicitKeywords, these distributions are specified via their quantiles, allowing for arbitrary selection with a small example shown in Figure~\ref{fig:quantile-pdf}. However, it is also possible to directly specify any desired distribution. In Appendix~\ref{app:sparsity_params}, we provide detailed specifications of the distributions utilized  to sample parameters for keyword initialization across all the experiments.

In the \env{} environment, the default approach for specifying parameter distributions is to utilize a list of quantiles. To clarify, when we mention ``sampling from quantiles,'' we are referring to the process of sampling from a distribution where the density function is determined by equal volume bins, with the left and right boundaries defined by each quantile.

The data format we adopt for each ``quantile'' is a triple specified by the user, representing a pair of bins encompassing minimum, median, and maximum values. An illustrative depiction of the resulting distribution can be found in Figure~\ref{fig:quantile-pdf}, showcasing an example of this approach.
\begin{figure}[htbp]
\centering
    \includegraphics[width=1.0\textwidth]{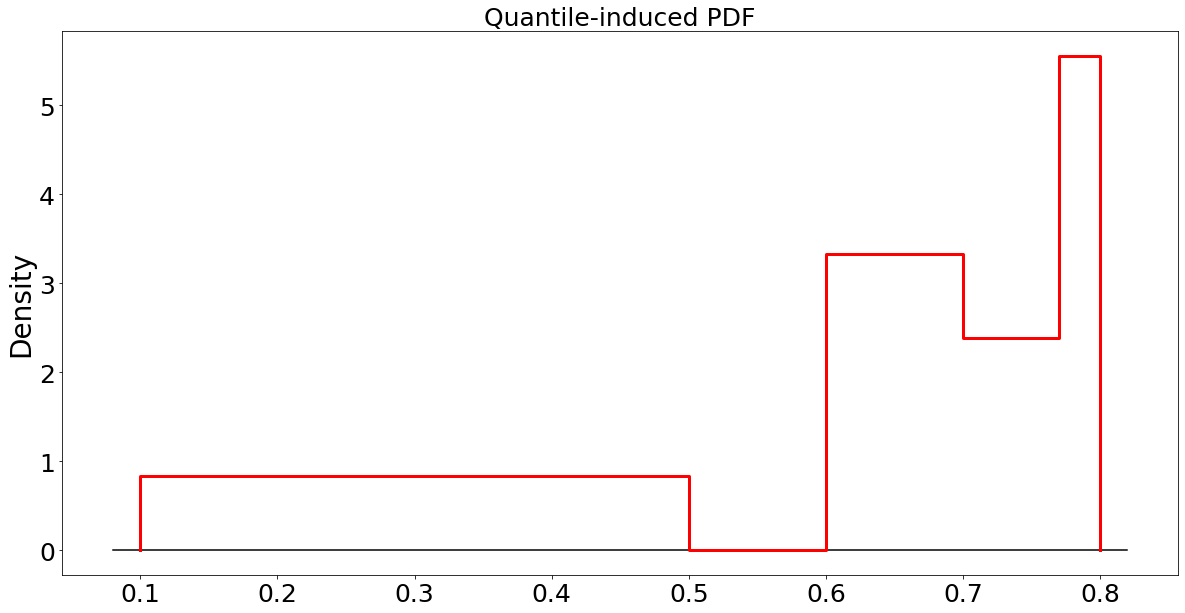}
    \caption{The probability density function (PDF) specified by the [min, median, max] quantile triples \\\texttt{[[0.1, 0.3, 0.5], [0.6,0.65,0.7], [0.7,0.77,0.8]].}} 
    \label{fig:quantile-pdf}
\end{figure}

Here we provide a comprehensive overview of the default parametric distributions employed within each keyword, along with the corresponding parameters obtained through sampling to initialize them. It is important to note that all distributions mentioned below can be customized to user-specified distributions, and the ones described here represent the default choices adopted for the experiments presented in this paper.

In the context of each keyword, the daily auction volume is sampled from a normal distribution characterized by a mean parameter and a standard deviation. The resulting volume is then clipped to ensure non-negativity and rounded to the nearest integer value representing the number of auctions. The mean parameter of the normal distribution for each keyword's volume is initialized by sampling from the quantile-induced distribution specified by the user. The standard deviation is set as $1+\alpha m$, where $m$ is the mean parameter, and $\alpha$ is uniformly sampled from the range $[0,\, 0.5]$.

Regarding revenue per conversion, each keyword samples from a normal distribution with a given mean and standard deviation, with the results clipped to a minimum of $0.01$ (one cent). The mean parameter for the revenue per conversion is sampled from the quantile-induced distribution specified by the user. The standard deviation is set as a proportion of the mean, and the proportion is sampled from the user-specified quantiles.

For the CTR and CVR parameters, which are single float values rather than distributions, their values are directly sampled from user-specified quantiles.

It is important to note that the keyword flow differs for ImplicitKeyword and ExplicitKeywords, as each of them incorporates additional internal distributions that are specific to their respective flows.

Implicit keywords incorporate internal distributions for the number of bidders and the distributions from which their bids are drawn. By default, when initializing keywords using quantiles, the number of competing bidders for an implicit keyword is deterministically set to a single competitor. This choice allows the bidding distribution to directly model the ``second price'' in second-price auctions. For each auction, the distribution used to sample the second price bid for a given keyword follows a Laplace distribution. To ensure non-negativity, the absolute value of the sample is used as the final bid. The location parameter of this distribution is sampled from the user-specified quantiles. The scale parameter is set as a proportion of the location parameter, with the proportion itself being sampled from quantiles.  As the price paid in a second-price auction corresponds to the bid of the bidder in second place, the sampled bid serves as the cost per click (CPC) when an auction is won.

The parameters sampled from quantiles to specify a set of ImplicitKeywords include the mean auction volume, location parameter for CPC, scale parameter for CPC (multiplier on location), CTR, CVR, mean revenue per conversion, and standard deviation of revenue per conversion (multiplier on mean).

Explicit keywords utilize internal functions that map bids to impression chance and bids to costs. By default, the bid to impression chance map follows a logistic function with a user-specified inflection point and tangent at the inflection point. The cost map, which is identical for all keywords by default, stochastically maps a bid to a normal distribution clipped between 0 and the bid. The mean of this distribution is set as a proportion of the bid, while the variance increases as the bid value increases. Although ExplicitKeywords are not utilized in the experiments presented in this paper, our framework includes functions to initialize sets of ExplicitKeywords. In the absence of user-specified quantiles, \env{} defaults to initializing ExplicitKeywords by sampling parameters from various beta or uniform distributions, which are then appropriately re-scaled.

Three additional parameters that impact keyword flow are the non-stationary stepsizes for mean volume, CTR, and CVR. These step sizes are uniform across all keywords but are scaled to the specific keyword's parameters. The step size for mean volume is multiplied by the initial mean volume, while the CTR and CVR steps act as multipliers on the keyword's current CTR and CVR values.

To introduce non-stationarity, a mask is utilized to designate each keyword as either stationary or non-stationary, allowing for precise control over the level of non-stationarity. In cases where all keywords are intended to remain stationary, passing a mask consisting of all False values is a viable option. Alternatively, passing None as the mask entirely skips the non-stationary computations, resulting in slightly improved computational efficiency.

In our experiments, all keywords are ImplicitKeywords. Experiments which in a stationary scenario involve setting all keywords to be stationary passing None as the mask, while the non-stationary experiments assign all keywords as non-stationary with a fixed step size of $\eta=0.03$ for all three variable parameters.
\section{Example of Bidding Outcomes}\label{app:bidding_example}
In this section, we explicitly demonstrate, using a two keyword example how the outputs are computed and what exactly they look like during two days of bidding.
Both keywords are initialized randomly with the same helper function that generated keywords for the sparsity heatmaps in Figure~\ref{fig:sparsity_heatmaps} with average volume set to 16 auctions per day and conversion rate set to 0.5. The first has a standard deviation of $1$ auction, and the second has a larger standard deviation of 8 auctions from the mean.
The bidding distributions are given by absolute value of Laplacian samples with location parameters approximately $0.65$ and $0.76$ and scale parameters of about $1/9$ and $1/5$ respectively. 
The first keyword has a CTR of about $0.75$, while the second has a CTR of about $0.1$. Both have a CVR of $0.5$, and their rounded Gaussian revenue per conversions are parameterized by ($1.23,\, 0.318$), and ($0.55,\, 0.09$).

Code to produce intermediate bidding outcomes such as those in the example below is provided in the \texttt{adcraft/appendix\_bidding\_outcomes\_example/} directory of the \env{} repository. 

Consider an agent that bids \$$0.75$ on each keyword to start with a budget of \$10. Consequently, its action is represented as:
{\small\begin{verbatim}
{budget: 10.0,  keyword_bids: [0.75, 0.75]}.
\end{verbatim}}
For the first keyword, a sample is drawn from the volume distribution to determine the number of auctions. A total of 15 is sampled, leading to the drawing of 15 bids from the critical bid distribution necessary to win an auction. These 15 critical bids are denoted as: 
\texttt{[0.67, 0.60,  0.62, 0.81, 0.56, 0.68, 0.46, 0.76, 0.74, 0.57, 0.79, 0.42, 0.60,  0.52,  0.63]}.
Impressions are received by the agent only for bids less than or equal to its standing bid of $0.75$. Out of the 15 bids, 12 meet this condition.Hence, the agent observes 12 impressions on the first keyword. 

For each of the 12 impressions with critical bids: 
\texttt{[0.67, 0.60,  0.62, 0.56, 0.68, 0.46, 0.74, 0.57, 0.42, 0.60,  0.52,  0.63]},
each impression is clicked with probability $0.7526828432972257$.  We draw a Bernoulli with that parameter for each impression and consider a click on each critical bid corresponding to positive outcomes.
From the 12 impressions, 10 positive outcomes are sampled, leading to 10 clicks with costs derived from the corresponding critical bids:
\texttt{[0.60, 0.62, 0.56, 0.68, 0.46, 0.74, 0.57, 0.42, 0.60, 0.63]}.

The agent, however, does not see these individual costs. It only observes the total number of clicked ads, which is 10, along with the combined cost of all these clicks for the first keyword: \$$5.88$. Subsequently, Bernoulli samples are drawn again to determine the number of conversions from these 10 clicks. The result is 4 conversions with individual revenues provided by samples from the keyword's revenue distribution:
\texttt{[1.53, 1.25, 1.87, 1.29]}. Again, the agent does not view these individual revenues but observes that it secured 4 conversions with a total value of $5.94$. Thus, the agent could compute the profit it earned on this keyword as \$$0.06$.

A similar process ensues for the second keyword. First, a volume sample of 12 is drawn, and 12 critical bids are sampled:
\texttt{[0.94, 0.72, 0.89, 0.87, 1.18, 0.44, 0.89, 1.17, 1.34, 0.01, 1.03, 1.45]}.
The agent's bid of $0.75$ on this keyword yields impressions for only 3 of these auctions, specifically those with bids \texttt{[0.72, 0.44, 0.01]}.
For each of those, a Bernoulli with parameter $0.10219080013611848$ samples which of those are clicked. Unluckily the agent receives no clicks. It observes 3 impressions, 0 clicks, $0.00$ cost, 0 conversions, and $0.00$ revenue on this keyword.
The observations that the agent receives for this day of bidding on the two keywords are given by the following dictionary, 
{\small\begin{verbatim}
{
  impressions:          [12,  3]
  buyside_clicks:       [10,  0]
  cost:                 [5.88,0.00]
  sellside_conversions: [10,  0]
  revenue:              [5.94,0.00]
  days_passed:           1
  cumulative_profit:     0.06
}
\end{verbatim}}
Perhaps the agent notices that on the first keyword it receives an average revenue of \$$0.594$ per click on the first keyword and lowers its bid to \$$0.59$. Not seeing any clicks or conversions, but observing a few impressions, perhaps the second keyword's bid is left at \$$0.75$ to collect more data.
The next action would then be
{\small\begin{verbatim}
{budget: 10.0,  keyword_bids: [0.59, 0.75]}.
\end{verbatim}}
To compute the outcomes for the first keyword, the volume is sampled. This time 16 auctions occur, and the critical bids sampled are\\
\texttt{[0.81, 0.76, 0.64, 0.57, 0.74, 0.84, 0.58, 0.65, 0.63, 0.85, 0.38, 0.71, 0.67, 0.34,
0.71, 0.28]}. Of those, 5 of them (\texttt{[0.57, 0.58, 0.38, 0.34, 0.28]}) are below the agent's bid, so that many impressions are observed. Clicks are sampled with a probability equal to the CTR of about 0.75. This time only two were clicked. The individual clicks' costs are given by the critical bids of \texttt{[0.58, 0.34]}.

The agent observes that they received 2 clicks with a total cost of  $0.92$.
Then conversions are sampled with a 0.5 probability. The lucky agent received a conversion on each click, with individual revenues of $[1.6, 1.48]$.
The agent observes that it received 2 conversions with a total revenue of $3.08$.

On the next keyword, a volume of 20 is sampled, with 20 corresponding critical bids sampled:
 \texttt{[0.78, 0.62, 0.6, 1.07, 0.59, 0.46, 0.64, 0.80, 0.78, 0.96, 0.78, 0.64, 0.72, 0.97,
0.82, 1.29, 0.69, 0.78, 0.8, 1.01]}.

Eight of those lie below the agent's bid of 0.75, leading to impressions with critical bids of
\texttt{[0.62, 0.60, 0.59, 0.46, 0.64, 0.64, 0.72, 0.69]}. The roughly 0.1 CTR is used to sample which impressions are clicked leading to 2 clicks with costs \texttt{[0.46 0.69]}. The agent observes 8 impressions, 2 clicks, and total cost 1.15.
Once again lucky, the 0.5 conversion rate led to two conversions, and the respective sampled revenues were \texttt{[0.52, 0.55]}. The agent observes 2 conversions and total revenue of $1.07$.
The full observation set for this next day of bidding is given by
{\small\begin{verbatim}
{
  impressions:          [5,   8]
  buyside_clicks:       [2,   2]
  cost:                 [0.92,1.15]
  sellside_conversions: [2,   2]
  revenue:              [3.08,1.07]
  days_passed:           2
  cumulative_profit:     2.14
}.
\end{verbatim}}
Here, the cumulative profit is the sum of the previous day's cumulative profit of \$$0.06$ with this new day's profit of \$$2.08$.
If the agent had used a smaller budget of, say, \$2 on the second day rather than \$10, then the final click on the second keyword with a corresponding critical bid of $0.69$ would have been too expensive to occur. In fact, that auction would not have been won at all because not enough budget would remain to pay for the cost if it were clicked. The resulting observations would be
{\small\begin{verbatim}
{
  impressions:          [5,   7]
  buyside_clicks:       [2,   1]
  cost:                 [0.92,0.46]
  sellside_conversions: [2,   1]
  revenue:              [3.08,0.52]
  days_passed:           2
  cumulative_profit:     2.28
}.
\end{verbatim}}
In that particular example, a lower budget prevents an expensive click leading to slightly more profit, though this is not always the case.
If the budget for the same samples were only \$$1.00$, then the interleaved sub-time steps would play a greater role. Both keywords would lose impressions. With the ratio of volumes, 4 impressions occur on the first keyword for every 5 on the second. Time-wise, the first clicked impression belongs to the second keyword with a $0.46$ cost. After that, any impression with a critical bid greater than $1.00-0.46=0.54$ would be eliminated. Therefore, the $0.58$ cost impression on the first keyword would be removed, and its corresponding Bernoulli sample leading to a click would occur on the next impression cheap enough to be observed at $0.38$ on the first keyword.
Afterward, all impressions would be too expensive for the budget.
If the revenues and samples remained the same, then the resulting observations for action \texttt{\{budget: 1.00,  keyword\_bids: [0.59, 0.75]\}} would be
{\small\begin{verbatim}
{
  impressions:          [2,   4]
  buyside_clicks:       [1,   1]
  cost:                 [0.38,0.46]
  sellside_conversions: [1,   1]
  revenue:              [1.6,0.52]
  days_passed:           2
  cumulative_profit:     1.28
}.
\end{verbatim}}
In practice, the revenue values and conversions may differ due to implementation. Revenues for each keyword are sampled using the same random generator as its Bernoulli samples for conversions and clicks. With fewer impressions, fewer samples are taken and so the current seed at the time of the samples would be different than in this example.
\section{Evaluation Metrics Formulations}\label{app:metric_formulations}

The profit associated with a given keyword $k$ is determined by subtracting the cost of all ad clicks for keyword $k$ from the sum of revenue obtained from conversions on $k$. The distributions from which these values are sampled, as well as the distributions and values influencing the number of clicks and conversions for each keyword at a given bid, exhibit variations across different keywords and instances of the \env{} environment. Consequently, directly comparing profits or other effectiveness measures for a bidder between instances of \env{} can be challenging. To facilitate more meaningful comparisons between instances, it is beneficial to normalize the achieved profit by some reference measure. Here, we formally define the Normalized Cumulative Profit (NCP) and Average per-Keyword Normalized Cumulative Profit (AKNCP) as specific realizations of normalized profit.

\textbf{Normalized Cumulative Profit (NCP):} 
In the NCP metric, we normalize the cumulative profit obtained by the agent by comparing it to the cumulative maximum expected profit that optimal bidding would yield with an infinite budget. This metric serves an indicator of the agent's effectiveness in optimizing total profit. The subsequent equation illustrates the mathematical formulation for the NCP metric, where profit$_{t,k}(b)$ is a random variable denoting profit (or loss) received for bidding $b$ on keyword $k$ in timestep $t$, and profit$_{t,k}$ denotes the actual profit of the agent during timestep $t$ for keyword $k$.
\begin{align} \label{eq:NCP}
    \text{NCP} 
    = \frac{\text{Total agent profit}}{\text{Max expected profit}} 
    = \frac{
        \displaystyle
        \sum_{k=0}^{K-1}
        \sum_{t=0}^{T-1}
            \mathrm{profit}_{t,k}
    }
    {
        \displaystyle
        \sum_{k=0}^{K-1}
        \sum_{t=0}^{T-1}
            \max_{b}
            \ex{
                \mathrm{profit}_{t,k}\paren{
                    b
                }
            }
    }
\end{align}
The computation of $\ex{\mathrm{profit}_{t,k}\paren{b_{t,k}}}$ is discussed in Section~\ref{appsec:opt_prof}
When no keywords are profitable on average, the maximum expected profit is 0, which would make the above equation for NCP undefined. Moreover, when optimal profit is actually a loss because bidding at all on a keyword leads to negative profit in expectation, the normalization becomes far less useful.  To curb those complications, when the optimal profit is $\leq0$ we don't perform any normalization, replacing the denominator with $1.0$.

\textbf{Average per-Keyword Normalized Cumulative Profit (AKNCP):} 
AKNCP averages the proportion of the maximum expected profit earned on each keyword. A high score on this metric implies that the agent bid effectively on the majority of keywords. Conversely, a low AKNCP score coupled with high NCP suggests that the agent primarily missed opportunities on unprofitable keywords or those with relatively lower profitability compared to the total profit achievable. The subsequent equation presents the mathematical formulation for AKNCP, where $\mathrm{Med}_k$ denotes the median value across all keywords. We use median in order to better explore the typical behavior of a keyword since the metric is unbounded below, and outliers can drastically skew the results when taking a mean.
\begin{align} \label{eq:AKNCP}
    \text{AKNCP} 
    &= \mathrm{Med}_{k} \left[ \frac{\text{total profit on keyword }k}{\text{Max expected keyword }k\text{profit}}\right]
    = \mathrm{Med}_{k} \left[
    \frac{
        \displaystyle
        \sum_{t=0}^{T-1}
            \mathrm{profit}_{t,k}
    }
    {
        \displaystyle\sum_{t=0}^{T-1}
        \max_{b}
        \ex{
            \mathrm{profit}_{t,k}\paren{
                b
            }
        }
    }
    \right]
\end{align}
For identical reasons to NCP, we replace each denominator with 1.0 to remove the normalization when a keyword has $\leq 0$ optimal profit. AKNCP assesses the agent's ability to bid optimally on individual keywords, considering expected profit per timestep. 

\subsection{Optimal Profit}\label{appsec:opt_prof}
To compute both the NCP and AKNCP, we normalize each keyword's profits by $\ex{\mathrm{profit}_{t,k}\paren{b}}$, which we call the optimal profit. While this measure may not be directly computable from observations from an agent's bids, it can be estimated through additional computations using the internal parameters and samples from the internal keyword distributions. Estimates of the mean values can be directly obtained for parameters such as mean volume, revenue per conversion, CTR, and CVR. However, samples from the internal distributions are necessary to determine the typical cost per click for each bid and the typical probability of winning an auction for each bid.

Specifically, for a given bid $b$ and a distribution $\mathcal{C}$ representing second price bids (costs per ad click), the expected cost of an ad click given an impression at bid $b$ is given by $\expect{c\sim\mathcal{C}}{c \,|\, c \leq b}$. Similarly, the impression chance for a given bid $b$ can be calculated by evaluating $\Pr(c\leq b\,|\,c\sim \mathcal{C}) = \expect{c\sim\mathcal{C}}{\indic_{c \leq b}(c)}$ where $\indic$ is the indicator function. Efficient computation of these quantities can be achieved by sampling a large set of second price bids $C_1, \dots, C_S$ where the required number of samples $S$ depends on the desired confidence interval for the convergence of the cumulative density function and quantiles.

The sorted list $C_1, \dots, C_S$ is used to determine the correct location of each bid within the list. Sorting the list requires $O(S\log S)$ time, while finding the correct indices for $B$ distinct bids takes $O(B\log S)$ time.
The proportion of samples less than each bid $b$ provides an estimate of the impression rate, and the mean of the samples less than $b$ represents the expected cost. Computing the means can be done in one pass by calculating the cumulative sum of the sorted samples in $O(S)$ time. For each bid $b$, the expected cost is obtained by dividing the cumulative sum at the index left of $b$ by the index. This requires $O(B)$ lookups after all the locations have been computed. Overall, estimating these quantities for every keyword requires an additional computation of $O(K(B+S)\log S)$.

To bound the number of bids $B$, it is important to note that bids can only be submitted in one cent increments. Additionally, a bid higher than the largest possible profit or the largest possible second price bid will not result in more profit. Therefore, the maximum allowed bid can be set to the higher of the mean cost per click plus a certain number of the largest possible scale multiples of that mean, or the largest value specified by a set of quantiles. In our experiments, the largest possible revenue mean was 1.5, and we used $B=300$ with bids ranging from $0.01$ to $3.00$ in one cent increments. We sampled $S = 2048$ from the cost distributions of each keyword. Even for non-stationary settings, $\mathcal{C}$ is fixed in time, so this computation needs to be performed only once before any bidding begins.

With these computations, we can determine the expected profit per bid for each keyword. As the distributions internal to a keyword are independent in our experiments, the expected profit from a bid $b$ for a given keyword can be factored as a product of expectations and scalars. Each expectation is taken over an internal distribution of the keyword, such as the revenue per click distribution. The distribution of competing bids on the keyword is denoted by $\mathcal{C}$.
\begin{align}
&\ex{\mathrm{profit}(b)} = \ex{\mathrm{\#\ clicks\ on\ keyword\ }k\mathrm{\ for\ bid}\ b} \cdot \ex{\mathrm{profit\ per\ click\ on\ keyword\ }k\mathrm{\ for\ bid}\ b}
\notag\\
&=
\big(\ex{\mathrm{\#\ auctions}} \cdot \expect{c\sim \mathcal{C}}{\indic_{c \leq b}(c)} \cdot \mathrm{CTR}
\big)
\cdot \big(
    \mathrm{CVR}
      \cdot  
      \ex{\mathrm{revenue\ per\ conv.}} 
    - \expect{c\sim \mathcal{C}}{c\,|\, c \leq b}
\big)
\notag\\
&=\ex{\mathrm{\#\ auctions}} \cdot \expect{c\sim \mathcal{C}}{\indic_{c \leq b}(c)} \cdot \Pr(\mathrm{ad click}\, |\, \mathrm{impression})
\notag\\
&\cdot \big(
    \Pr(\mathrm{conversion}\, |\, \mathrm{ad click}) 
      \cdot  
      \ex{\mathrm{revenue\ per\ conv.}} 
    - \expect{c\sim \mathcal{C}}{c\,|\, c \leq b}
\big)
\label{eq:profit}
\end{align}
For stationary keywords, the expected profits per bid can be computed once and remain fixed over time. However, for non-stationary keywords, only the portions of Equation~\ref{eq:profit} that depend on the bid need to be estimated once before any bidding begins. The remaining parts of the equation can be evaluated at each time step using the current mean volume, CVR, CTR, and mean revenue per conversion, which can be directly read from the keyword's internal parameters without the need for sampling. It should be noted that for volume and revenue, these values may slightly underestimate the true mean volume due to the minimum value clipping. However, in our experiments, the standard deviations are small enough that this has minimal effect. In general, volume and revenue can be computed using samples, with non-stationary volume requiring resampling at each time step while revenue per conversion and stationary volume require sampling only once.

The computation in Equation~\ref{eq:profit} can be vectorized, resulting in an array of expected profit per bid. During each time step, the optimal profit for each keyword is $\max_{b}\mathrm{profit}(b)$ which is the maximum value in the array.

\section{Baseline Bidder Details}\label{app:baseline_details}
The baseline bidder maintains a running average of the observed revenue per conversion and the probability of conversion given an ad click for each keyword. By combining these estimates, the bidder obtains an approximation of the expected revenue per ad click on each keyword. At the start, the agent lacks knowledge of the conversion rate and the revenue per conversion, so it initiates bidding at a fixed initial value. Until an ad click is observed, the agent incrementally increases its bids at each time step. Once an ad click is observed, the agent can begin estimating the conversion rate. If a conversion occurs, the agent also obtains an estimate of the average revenue per conversion. In cases where no conversion is observed, a default value is assumed for the revenue per conversion. Subsequently, the baseline agent bids its perceived value of the revenue per ad click, with the probability of bidding its true belief increasing as it receives more ad click observations.

Pseudocode detailing the behavior of the baseline agent is provided in Algorithm~\ref{alg:baseline-code}.

\begin{algorithm}[htbp]
\SetAlgoLined

\KwResult{Updated cache}
\textbf{Hyperparameters:}\\
\Indp
  default\_revenue\_per\_conversion $\leftarrow$ 1.0\\
  bid\_step $\leftarrow$ 0.03\\
  initial\_bids $\leftarrow$ 0.1 for each keyword\\
\Indm
\textbf{Cached Memory:}\\
\Indp
  revenue\_per\_conversion\_estimates $\leftarrow$ \{\}\\
  number\_of\_revenue\_observations $\leftarrow$ \{\}\\
  conversion\_rate\_estimates $\leftarrow$ \{\}\\
  number\_of\_conversion\_rate\_observations $\leftarrow$ \{\}\\
\Indm
\textbf{Other Variables:}\\
\Indp
  max\_bid\_from\_steps $\leftarrow$ \{initial\_bid[k] for each keyword k\}\\
\Indm
\textbf{Cache updates on each observation:}\\
\For{each keyword k}{
  \If{click(s) observed}{
    update  conversion\_rate\_estimates[k],  number\_of\_conversion\_rate\_observations[k]
  }
  \If{conversion(s) observed}{
    update  revenue\_per\_conversion\_estimates[k],   number\_of\_revenue\_observations[k]
  }
}
\textbf{Sample actions with cache:}\\
\For{each keyword k}{
  Sample $s\sim$ Uniform[0,1]\\
  \eIf{$s \leq 1 /$ number\_of\_conversion\_rate\_observations[k]}{
    bid[k] $\leftarrow$ max\_bid\_from\_steps + bid\_step
  }{
    \eIf{number\_of\_revenue\_observations[k] $= 0$}{
      bid[k] $\leftarrow$ conversion\_rate\_estimates[k] $\times$ default\_revenue\_per\_conversion
    }{
      bid[k] $\leftarrow$ conversion\_rate\_estimates[k] $\times$ revenue\_per\_conversion\_estimates[k]
    }
  }
}
\caption{Baseline bidder's pseudocode}\label{alg:baseline-code}
\end{algorithm}
\section{Sparsity Regimes} \label{app:sparsity_params}
\subsection{Parameters defining experimental sparsity regimes}
In this section, we outline the specific quantiles utilized for both the sparsity heatmap experiments and RL training experiments.

As detailed in Appendix~\ref{app:env_params}, the parameters sampled from quantiles to define a set of ImplicitKeywords include auction volume (mean), cost per click (location parameter), cost per click (scale parameter, a multiplier on location), CTR, CVR, revenue per conversion (mean), and revenue per conversion standard deviation (multiplier on mean).

Even limiting to a set of parametric distributions, there is still a significant level of flexibility in defining the environment. In our exploration, we focus on a straightforward set of quantiles where each parameter is sampled from a single pair of uniform bins with equal volume. This simplified setting allows us to examine the impact of fixing different choices of auction sparsity through mean volume, cost sparsity through CTR, and revenue sparsity through CVR. The specific set of quantiles adjusted for each experiment is provided in Table~\ref{table:experiment_generic_quantiles}. They are chosen both to mirror typical values one might see in real data, as well as to ensure that some portion of keywords are likely to be profitable. This allows us to determine how well models did at identifying profitability in each case. Further, the limit of standard deviation size eases the computation of optimal profits needed to compute our experimental metrics.

\begin{table}[h!]
\centering
\begin{tabular}{|l|c|c|c|} 
    \hline
    \textbf{Parameter} & \textbf{Minimum} & \textbf{Median} & \textbf{Maximum}\\
    \hline
    Mean auction volume & $64$ & $128$ & $256$\\ 
    \hline
    Cost per ad click Laplace location & $0.30$ & $0.55$ & $1.00$\\ 
    \hline
    Cost per ad click Laplace scale / location & $0.01$ & $0.15$ & $0.3$\\ 
    \hline
    $\Pr($click | impression) & $0.1$ & $0.5$ & $0.9$\\ 
    \hline
    $\Pr($conversion | click) & $0.1$ & $0.5$ & $0.9$\\ 
    \hline
    Mean revenue per conversion & $0.30$ & $1.0$ & $1.5$\\ 
    \hline
    Standard deviation revenue / mean revenue  &  $0.01$ & $0.15$ & $0.3$\\  
    \hline
\end{tabular}
\caption{Generic simple parameter quantiles. The associated distributions are defined as uniformly sampling from [Minimum, Median] with a probability of 0.5, and uniformly sampling from [Median, Maximum] otherwise. Each of our experiments is defined by these quantiles, with a pair of rows from this table modified to consistently yield a fixed value.}
\label{table:experiment_generic_quantiles}
\end{table}

We investigated the sparsity options within the \env{} environment by fixing pairs of sparsity parameters to specific values. An example of this is presented in Table~\ref{table:vol_cvr_quantiles}, where mean volume ($m$) and conversion rate ($p$) are set deterministically. These quantiles were used in experiments involving RL models, specifically in the "Dense regime" and "Sparse regime" scenarios. In the dense regime RL experiments, we sampled 100 keywords from the quantiles in Table~\ref{table:vol_cvr_quantiles} with $m=128$ and $p=0.8$. These settings are representative of real advertisements that are likely to result in conversions, such as links to articles that monetize via display ads shown to any user without ad blockers. In the sparse regime RL experiments, we sampled 100 keywords from the quantiles in Table~\ref{table:vol_cvr_quantiles} with $m=16$ and $p=0.1$. In real-world advertising, extremely sparse keywords may be searched for less than once per month or year. Achieving profitability through advertising to such keywords often requires prior knowledge of their monetization value and scaling advertising efforts to encompass a large number of such keywords.

However, our focus in this work is on learning the value of each advertisement. Therefore, we chose a regime with relatively higher traffic while still maintaining sparsity. By considering an average of $16$ auctions per day, we aim to capture the characteristics of keywords that are regularly searched for but remain unpopular. Effective monetization of many such keywords can lead to significant profitability. We chose a conversion rate of $0.1$ due to its placement left of the difficulty phase shift for the baseline model. However, in many common advertising scenarios, users who have already clicked on an advertisement are more likely to convert, making $0.1$ a somewhat conservative choice of conversion rate. While this sparse regime presents challenges for the baseline model, it represents potentially lucrative advertising scenarios. Developing a successful RL model capable of navigating these regimes could prove highly advantageous in real-world advertising scenarios. The consideration of these factors influenced our decision in selecting these specific sparsity parameters.

\begin{table}[h!]
\centering
\begin{tabular}{|l|c|c|c|} 
    \hline
    \textbf{Parameter} & \textbf{Minimum} & \textbf{Median} & \textbf{Maximum}\\
    \hline
    Mean auction volume & $m$ & $m$ & $m$\\ 
    \hline
    Cost per ad click Laplace location & $0.30$ & $0.55$ & $1.00$\\ 
    \hline
    Cost per ad click Laplace scale / location & $0.01$ & $0.15$ & $0.3$\\ 
    \hline
    $\Pr($click | impression) & $0.1$ & $0.5$ & $0.9$\\ 
    \hline
    $\Pr($conversion | click) & $p$ & $p$ & $p$\\ 
    \hline
    Mean revenue per conversion & $0.30$ & $1.0$ & $1.5$\\ 
    \hline
    Standard deviation revenue / mean revenue  &  $0.01$ & $0.15$ & $0.3$\\  
    \hline
\end{tabular}
\caption{The quantiles used for sparsity experiments, where the mean volume parameter is fixed at $m$ for every keyword and the CVR is fixed at $p$ for every keyword, are presented in Figure~\ref{fig:sparsity_heatmaps}. Each cell in the figure represents the results obtained by running the baseline bidder on keywords sampled from these quantiles, with a specific choice of $m$ and $p$.}
\label{table:vol_cvr_quantiles}
\end{table}

\subsection{Sparsity regime heatmaps: minimum, median, and maximum baseline performance}
\begin{figure}[htbp]
    \begin{center}
    \includegraphics[height=0.265\textheight]{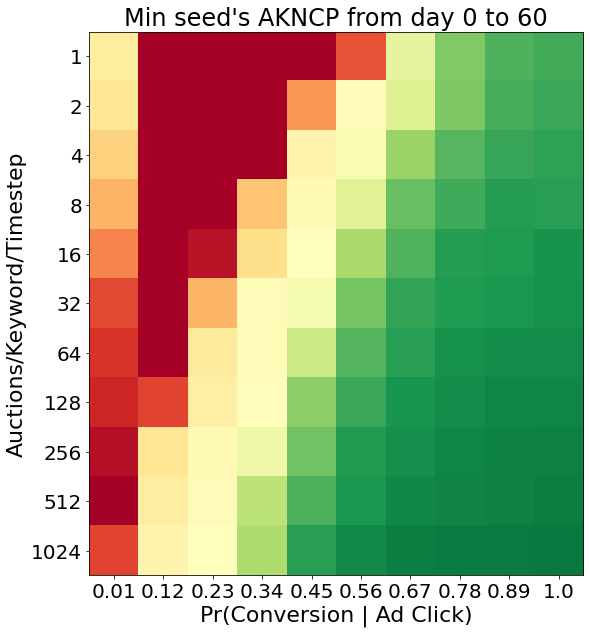} 
    \includegraphics[height=0.265\textheight]{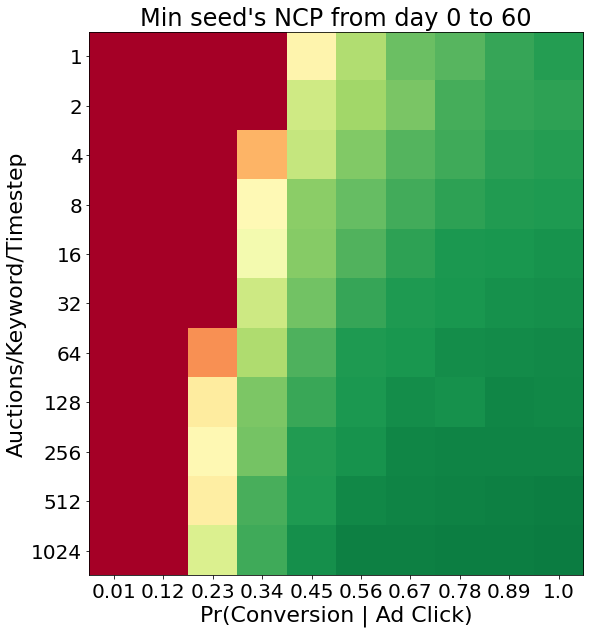}\\
    \includegraphics[height=0.265\textheight]{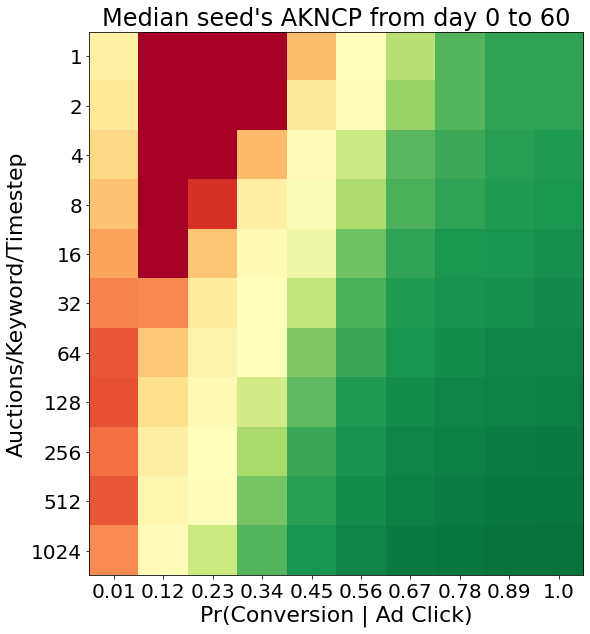} 
    \includegraphics[height=0.265\textheight]{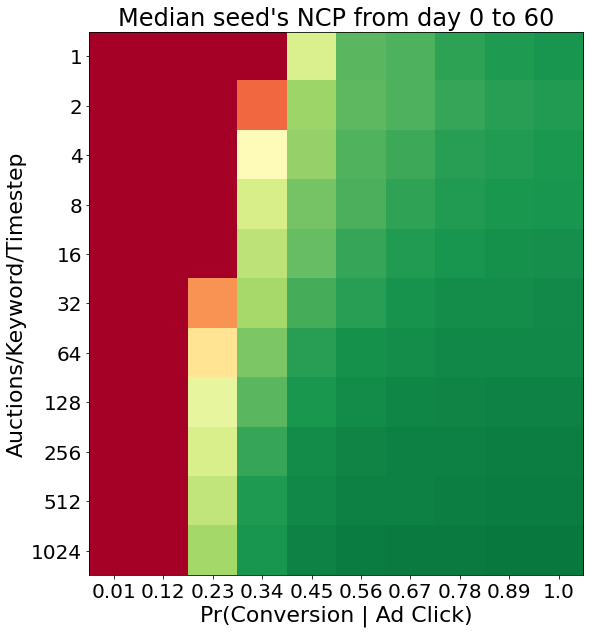} \\
    \includegraphics[height=0.265\textheight]{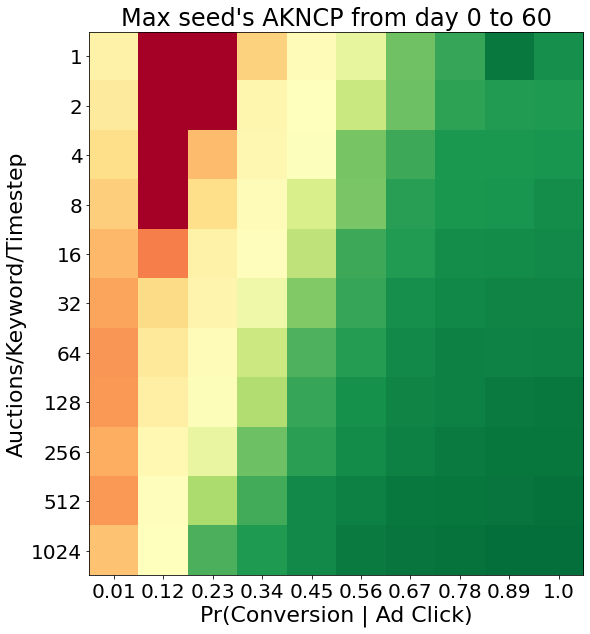} 
    \includegraphics[height=0.265\textheight]{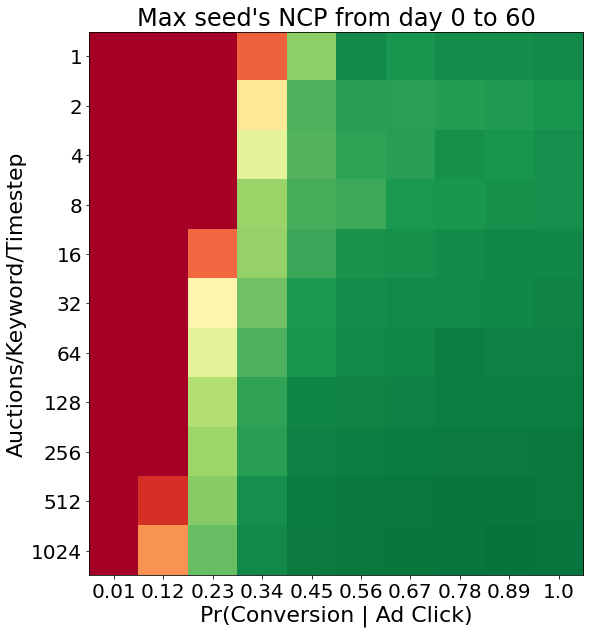} \\
    \hspace{2em}\includegraphics[width=0.8\textwidth]{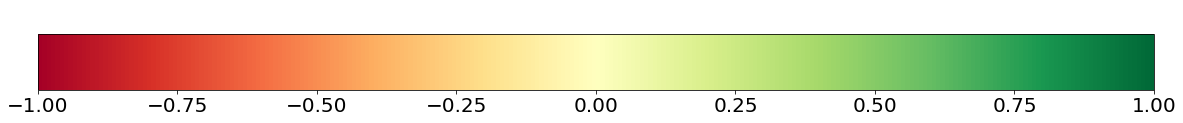}
    \end{center}
    \caption{Six signed heatmaps are presented, with rows representing fixed volume and columns representing fixed CVR. The left side of the heatmaps depicts the worst-, average-, and best-case performance of the baseline bidder in terms of AKNCP, while the right side shows the corresponding NCP performance. Scores below -1 are truncated to -1. The top row displays the worst seed performance for each cell's AKNCP and NCP, while the bottom row shows the best seed performance. The results in each cell represent the minimum, median, and maximum scores over 16 seeds. The medians are also reported in Figure~\ref{fig:sparsity_heatmaps}.}
        \label{fig:minmax_vol_sctr_heatmaps}
\end{figure}

\begin{figure}[htbp]
    \begin{center}
    \includegraphics[height=0.265\textheight]{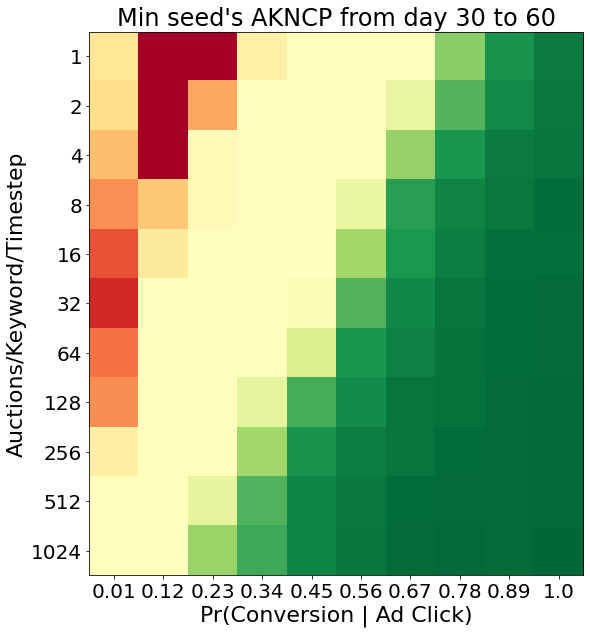} 
    \includegraphics[height=0.265\textheight]{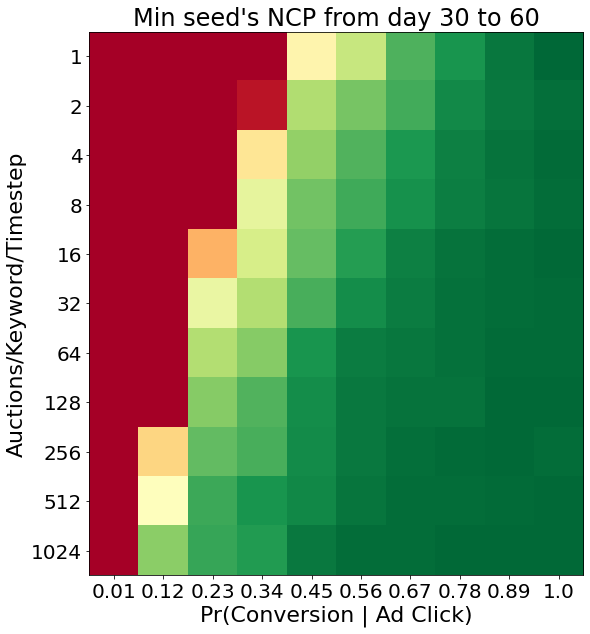}\\
    \includegraphics[height=0.265\textheight]{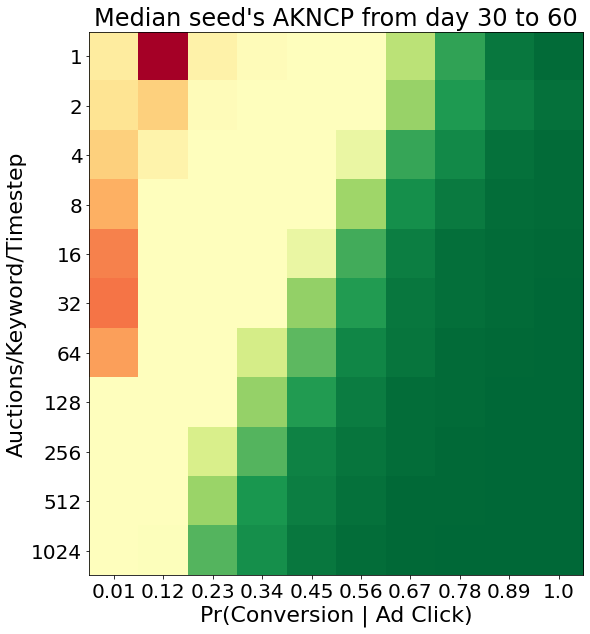} 
    \includegraphics[height=0.265\textheight]{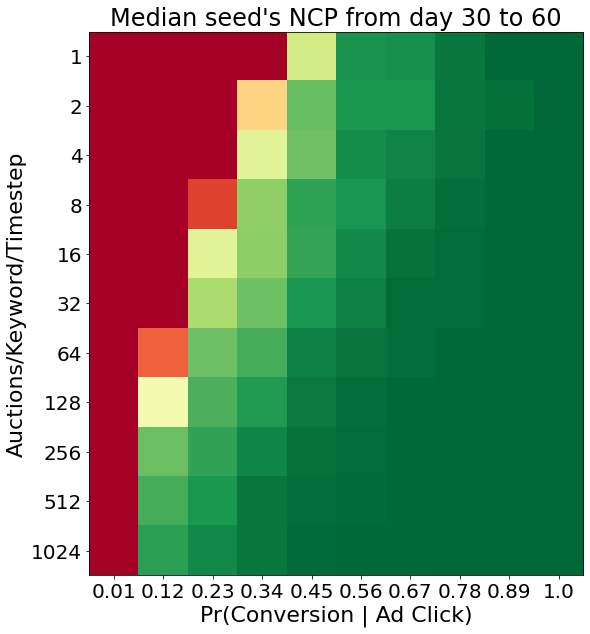} \\
    \includegraphics[height=0.265\textheight]{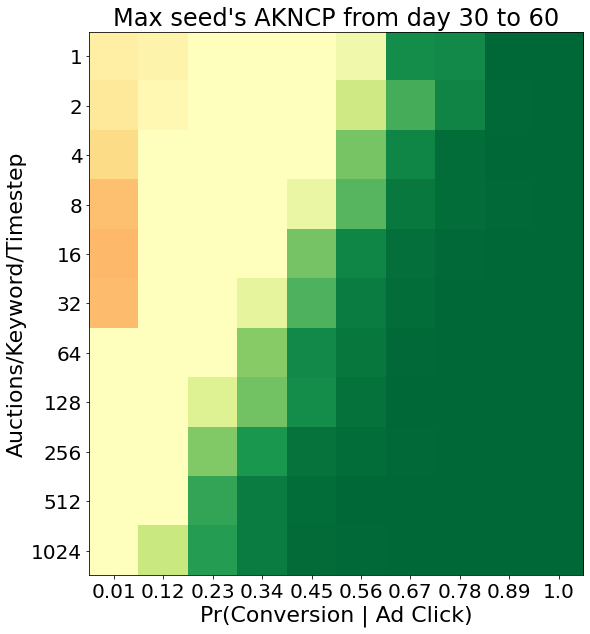} 
    \includegraphics[height=0.265\textheight]{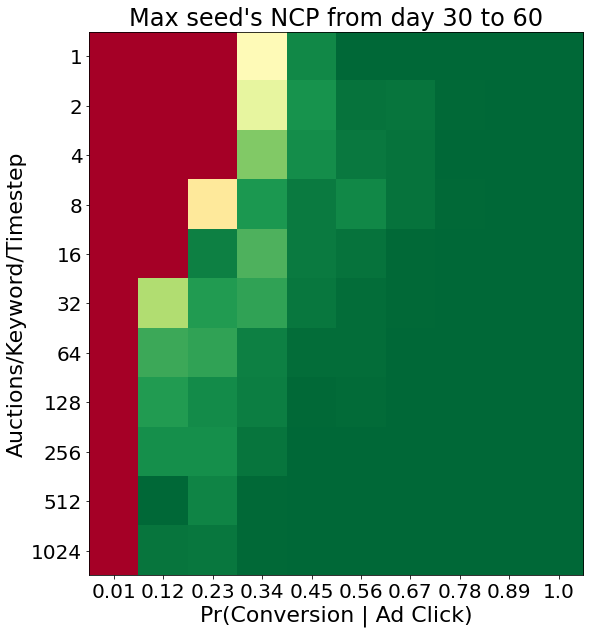} \\
    \hspace{2em}\includegraphics[width=0.8\textwidth]{heatmap_colorbar.png}
    \end{center}
    \caption{Six signed heatmaps are presented, with rows representing fixed volume and columns representing fixed CVR. The left side of the heatmaps depicts the worst-, average-, and best-case performance of the baseline bidder in terms of AKNCP after 30 days, while the right side shows the corresponding NCP performance. Scores below -1 are truncated to -1. The top row displays the worst seed performance for each cell's AKNCP and NCP, while the bottom row shows the best seed performance. The results in each cell represent the minimum, median, and maximum scores over 16 seeds. The medians are also reported in Figure~\ref{fig:sparsity_heatmaps}.}
        \label{fig:minmax_vol_sctr_heatmaps30}
\end{figure}

\begin{figure}[htbp]
    \begin{center}
    \includegraphics[height=0.265\textheight]{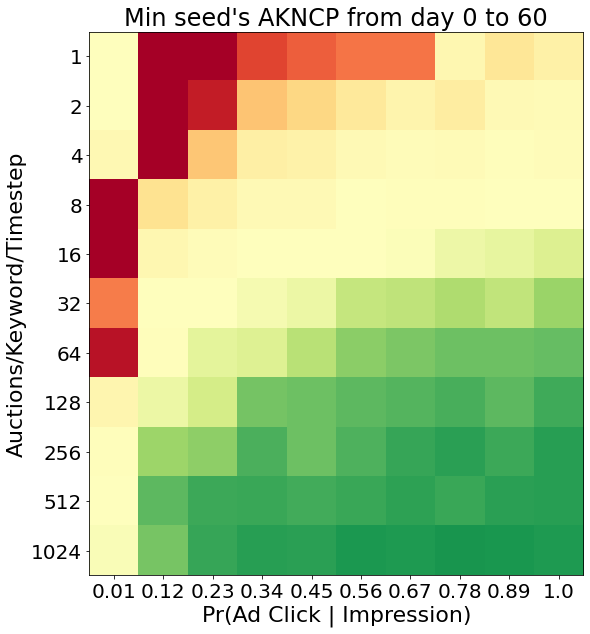} 
    \includegraphics[height=0.265\textheight]{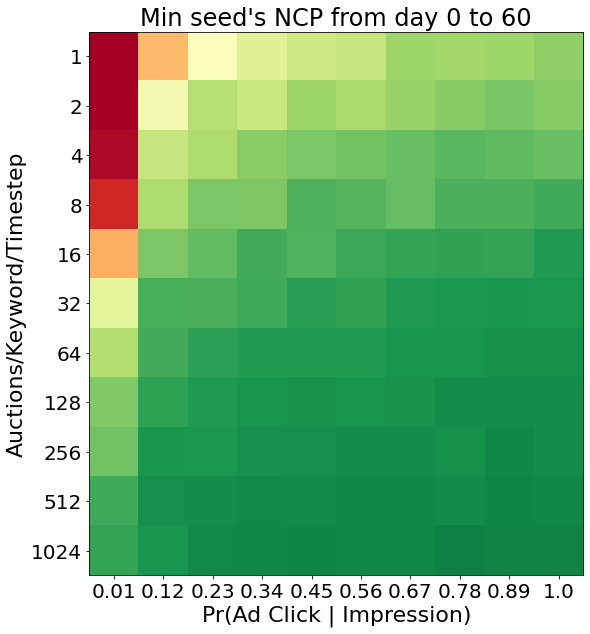}\\
    \includegraphics[height=0.265\textheight]{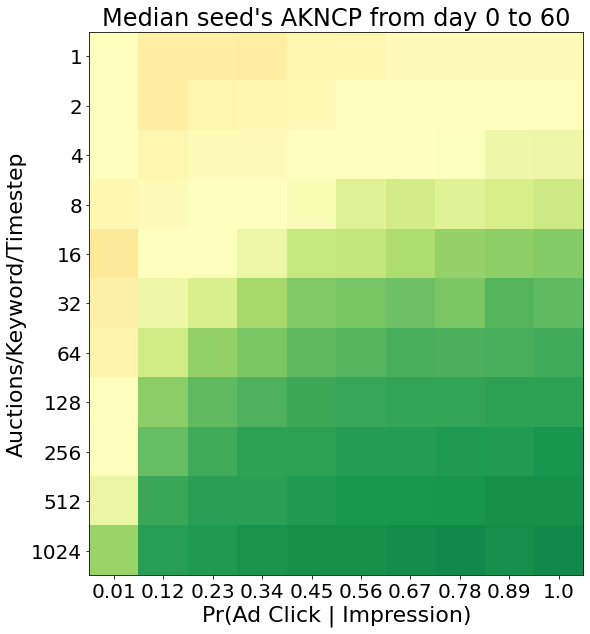} 
    \includegraphics[height=0.265\textheight]{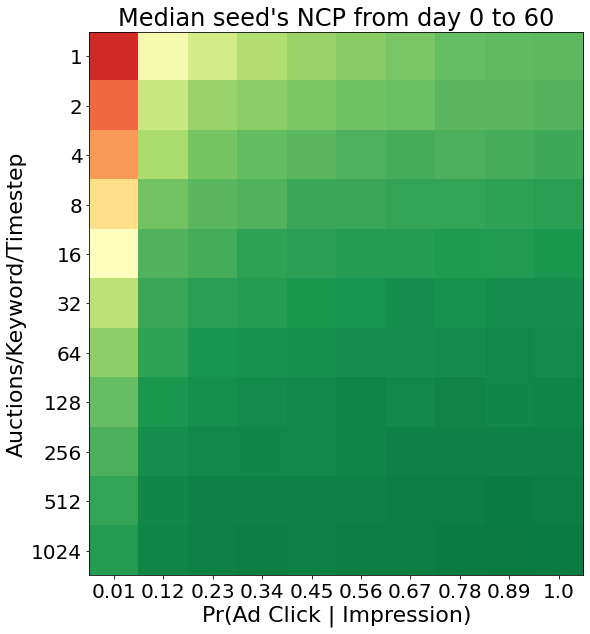} \\
    \includegraphics[height=0.265\textheight]{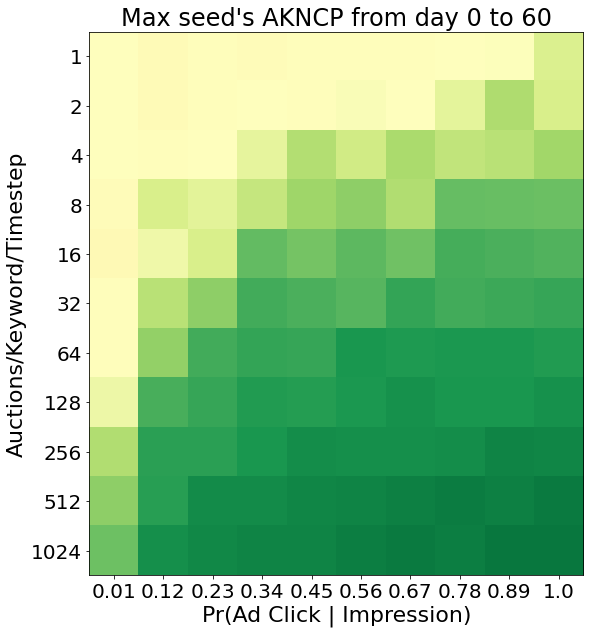} 
    \includegraphics[height=0.265\textheight]{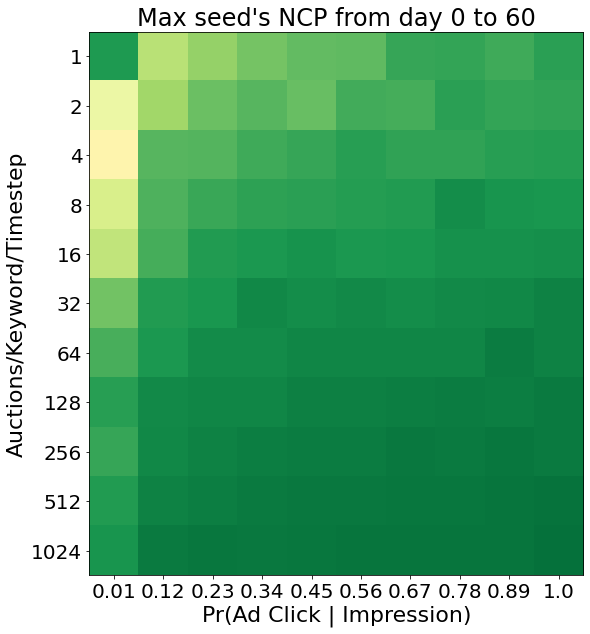} \\
    \hspace{2em}\includegraphics[width=0.8\textwidth]{heatmap_colorbar.png}
    \end{center}
    \caption{Six signed heatmaps are presented, with rows representing fixed volume and columns representing fixed CTR. The left side of the heatmaps depicts the worst-, average-, and best-case performance of the baseline bidder in terms of AKNCP, while the right side shows the corresponding NCP performance. Scores below -1 are truncated to -1. The top row displays the worst seed performance for each cell's AKNCP and NCP, while the bottom row shows the best seed performance. The results in each cell represent the minimum, median, and maximum scores over 16 seeds.}
        \label{fig:minmax_vol_bctr_heatmaps}
\end{figure}

\begin{figure}[htbp]
    \begin{center}
    \includegraphics[height=0.265\textheight]{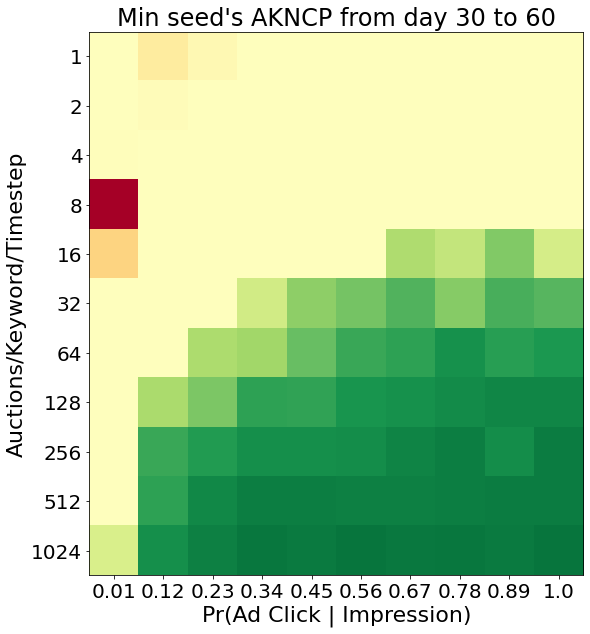} 
    \includegraphics[height=0.265\textheight]{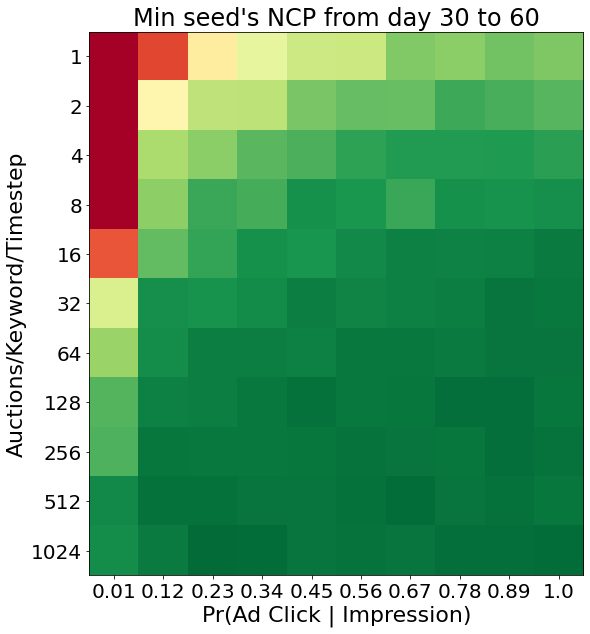}\\
    \includegraphics[height=0.265\textheight]{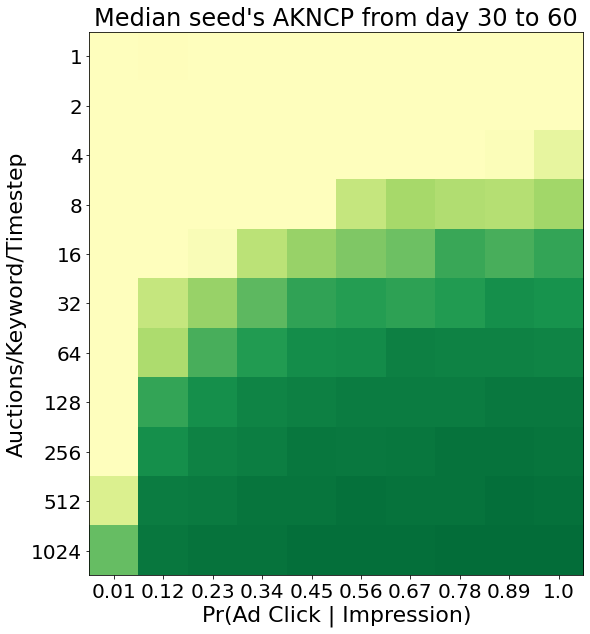} 
    \includegraphics[height=0.265\textheight]{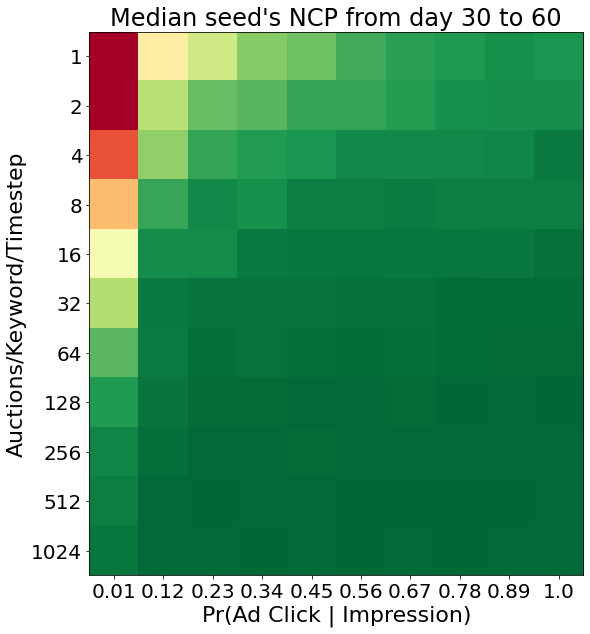} \\
    \includegraphics[height=0.265\textheight]{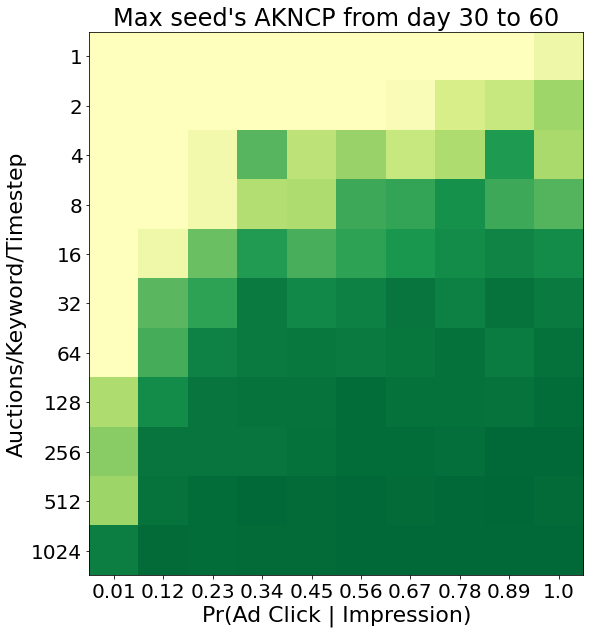} 
    \includegraphics[height=0.265\textheight]{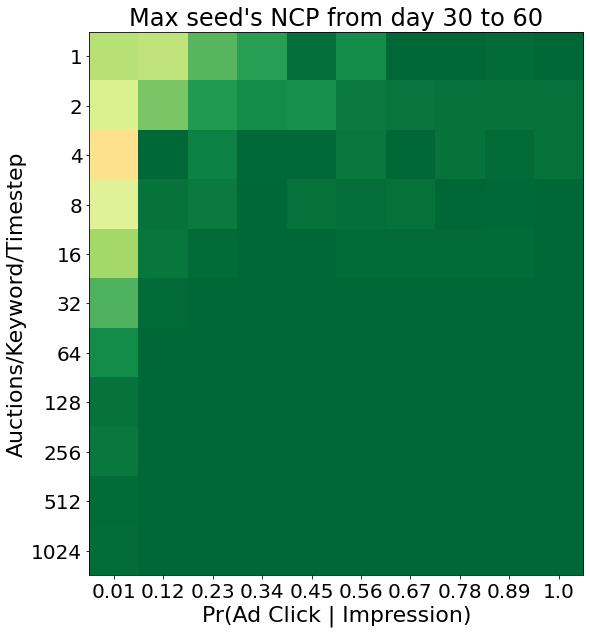} \\
    \hspace{2em}\includegraphics[width=0.8\textwidth]{heatmap_colorbar.png}
    \end{center}
    \caption{Six signed heatmaps are displayed, where the rows represent fixed volume and the columns represent fixed CTR. The left side of the heatmaps shows the worst-, average-, and best-case performance of the baseline bidder in terms of AKNCP after 30 days, while the right side depicts the corresponding NCP performance. Scores below -1 are truncated to -1. The top row represents the worst seed performance for each cell's AKNCP and NCP, while the bottom row displays the best seed performance. The results in each cell reflect the minimum, median, and maximum scores obtained from 16 seeds.}
        \label{fig:minmax_vol_bctr_heatmaps30}
\end{figure}

\begin{figure}[htbp]
    \begin{center}
    \includegraphics[height=0.265\textheight]{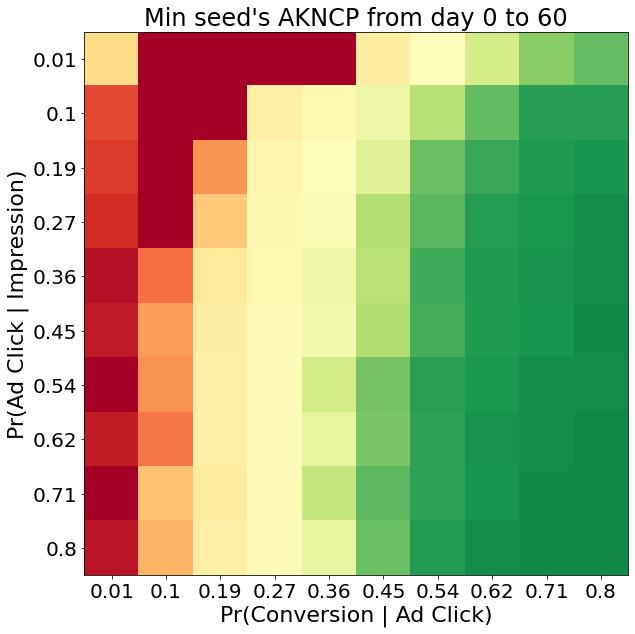} 
    \includegraphics[height=0.265\textheight]{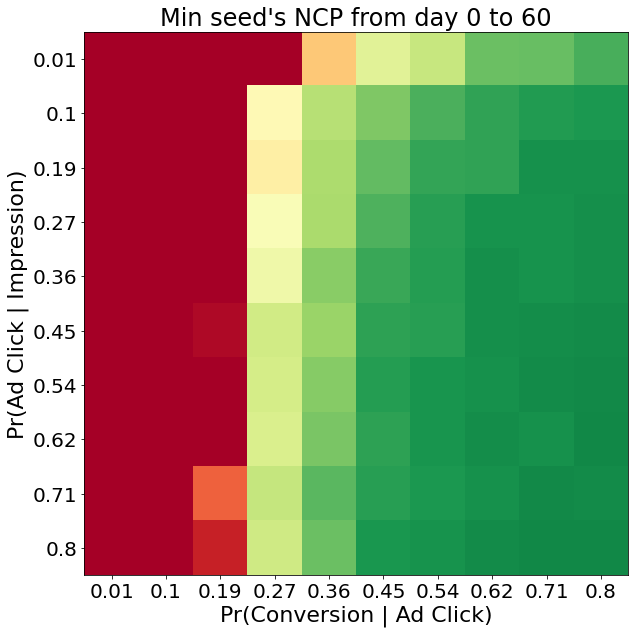}\\
    \includegraphics[height=0.265\textheight]{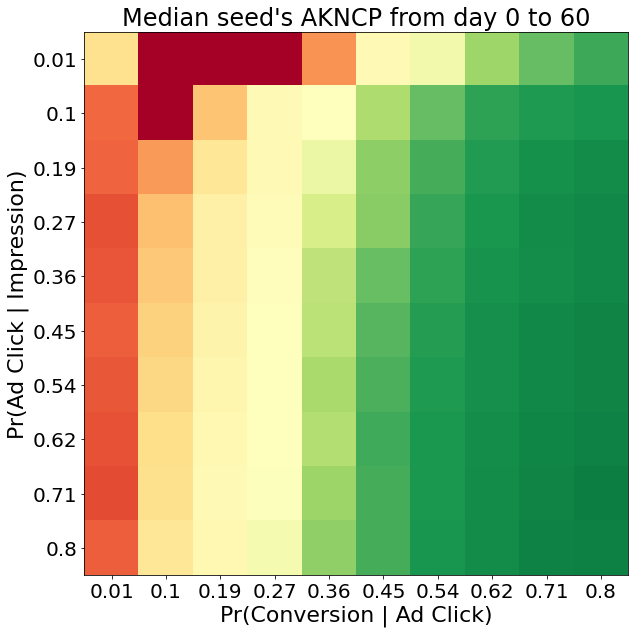} 
    \includegraphics[height=0.265\textheight]{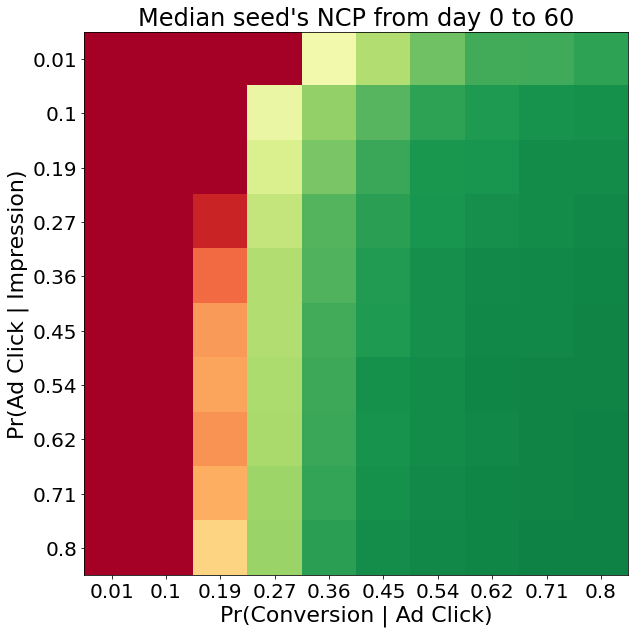} \\
    \includegraphics[height=0.265\textheight]{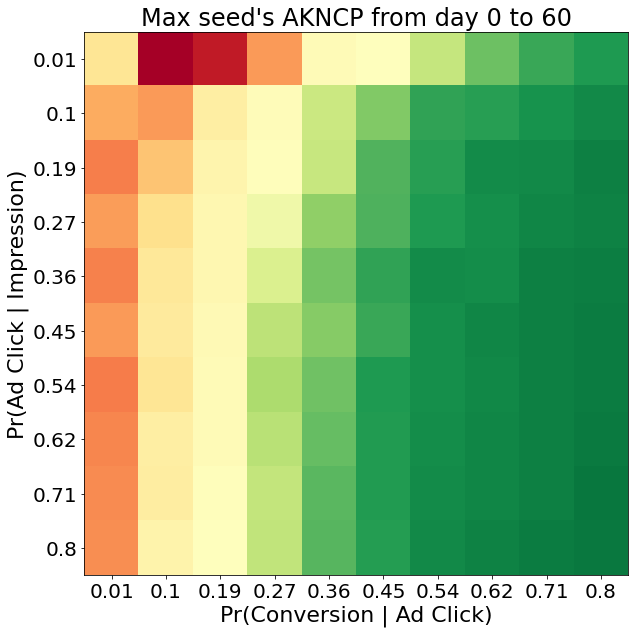} 
    \includegraphics[height=0.265\textheight]{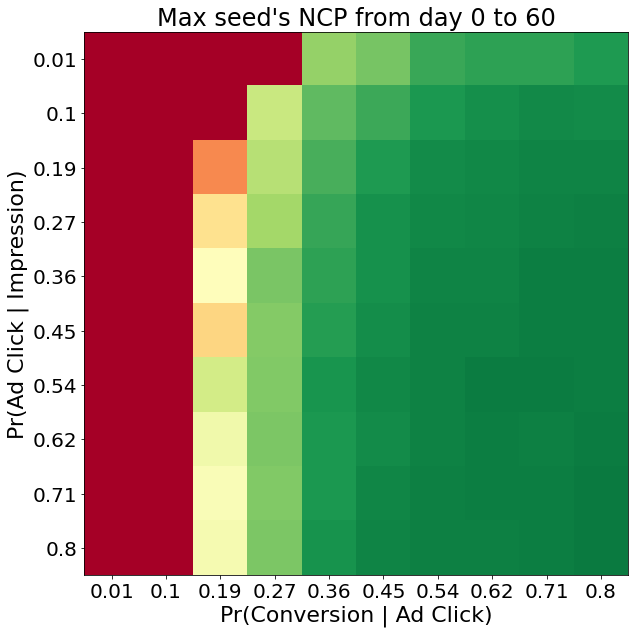} \\
    \hspace{2em}\includegraphics[width=0.8\textwidth]{heatmap_colorbar.png}
    \end{center}
    \caption{Six signed heatmaps are presented, with the rows representing fixed CTR and the columns representing fixed CVR. The left side of the heatmaps displays the worst-, average-, and best-case performance of the baseline bidder in terms of AKNCP, while the right side shows the corresponding NCP performance. Scores below -1 are truncated to -1. The top row represents the worst seed performance for each cell's AKNCP and NCP, while the bottom row displays the best seed performance. The results in each cell represent the minimum, median, and maximum scores obtained from 16 seeds.}
        \label{fig:minmax_bctr_sctr_heatmaps}
\end{figure}

\begin{figure}[htbp]
    \begin{center}
    \includegraphics[height=0.265\textheight]{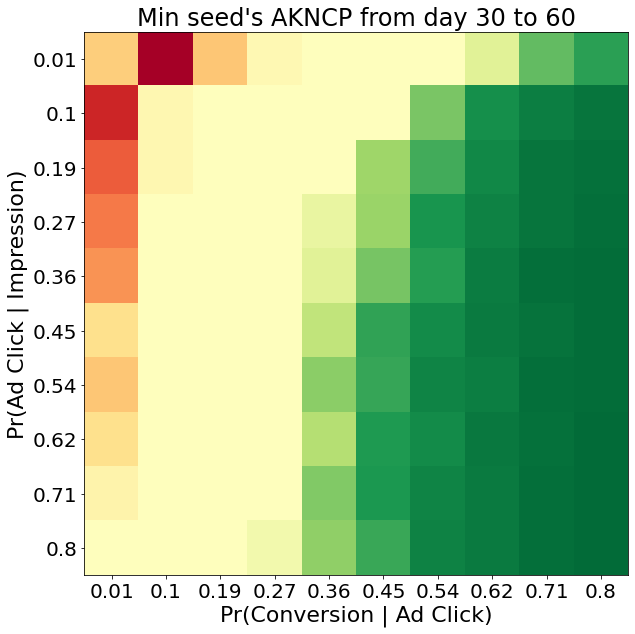} 
    \includegraphics[height=0.265\textheight]{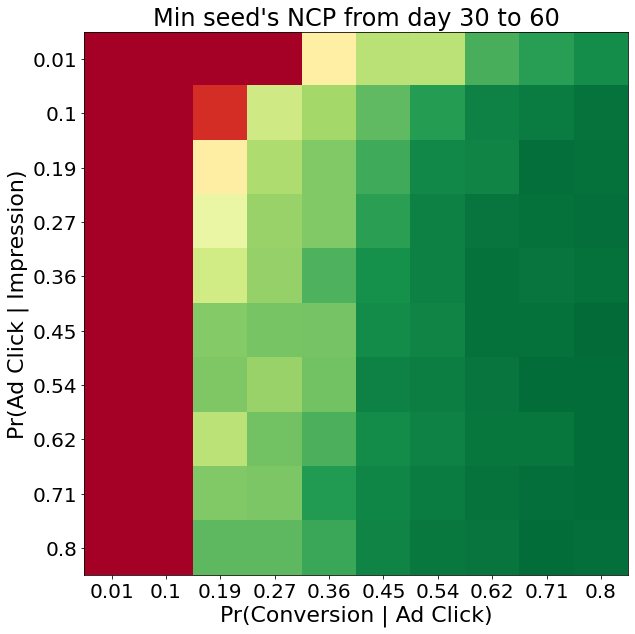}\\
    \includegraphics[height=0.265\textheight]{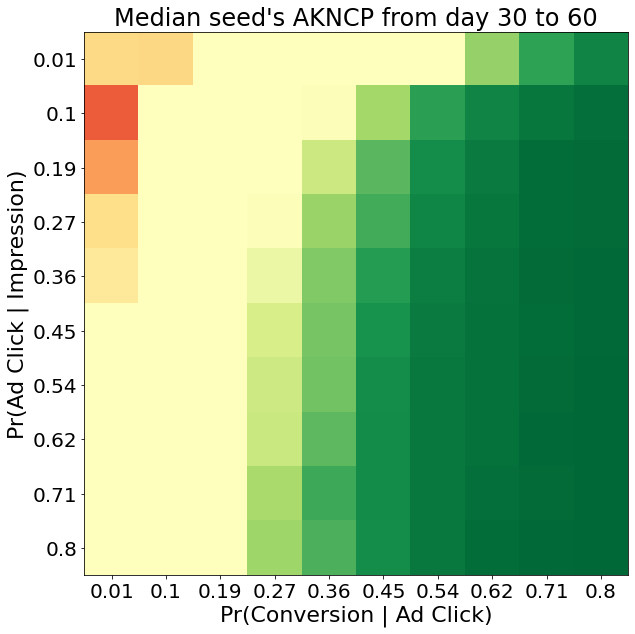} 
    \includegraphics[height=0.265\textheight]{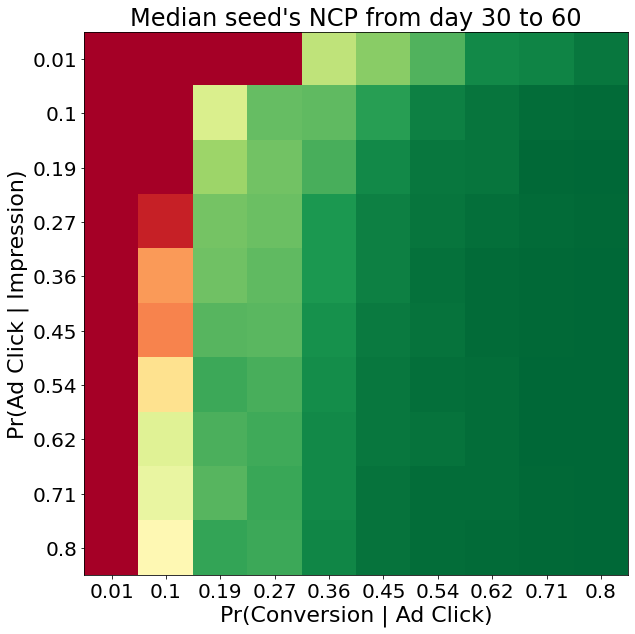} \\
    \includegraphics[height=0.265\textheight]{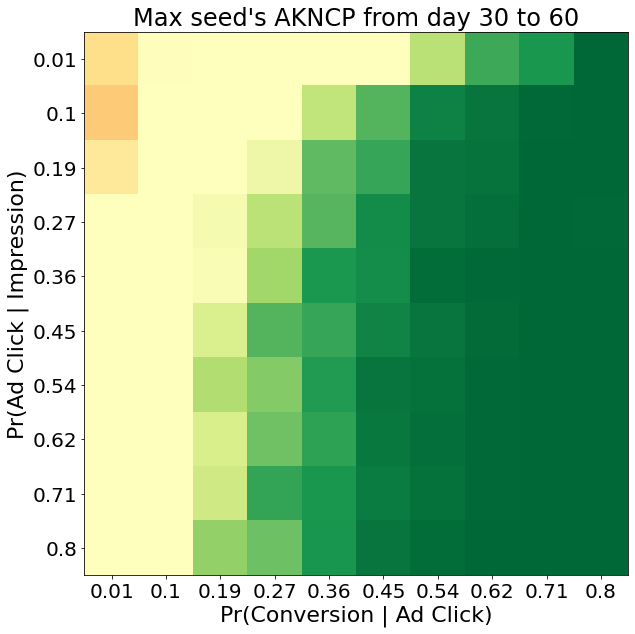} 
    \includegraphics[height=0.265\textheight]{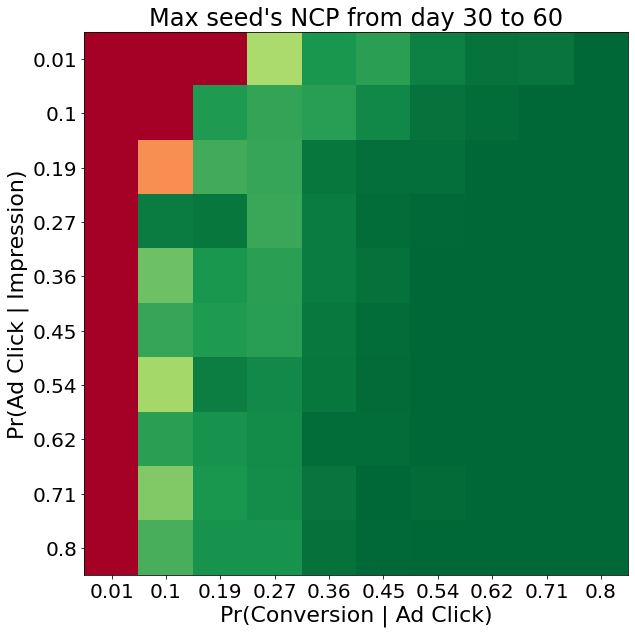} \\
    \hspace{2em}\includegraphics[width=0.8\textwidth]{heatmap_colorbar.png}
    \end{center}
    \caption{Six signed heatmaps are displayed, where the rows represent fixed CTR and the columns represent fixed CVR. The left side of the heatmaps shows the worst-, average-, and best-case performance of the baseline bidder in terms of AKNCP, while the right side depicts the corresponding NCP performance after 30 days. Scores below -1 are truncated to -1. The top row represents the worst seed performance for each cell's AKNCP and NCP, while the bottom row displays the best seed performance. The results in each cell represent the minimum, median, and maximum scores obtained from 16 seeds.}
        \label{fig:minmax_bctr_sctr_heatmaps30}
\end{figure}

In this section, we present the heatmaps illustrating the baseline performance on environments with 100 keywords sampled from quantiles for each pair of fixed sparsity parameters. Figures~\ref{fig:minmax_vol_sctr_heatmaps}and\ref{fig:minmax_vol_sctr_heatmaps30} display the performance on quantiles with fixed volume and conversion rate, as described in Table~\ref{table:vol_cvr_quantiles}. The reported median performances correspond to the same performances depicted in Figure~\ref{fig:sparsity_heatmaps}. Additionally, Figures~\ref{fig:minmax_vol_bctr_heatmaps}and\ref{fig:minmax_vol_bctr_heatmaps30} demonstrate the performance of the baseline bidder on quantiles specified in Table\ref{table:experiment_generic_quantiles}, but with fixed numbers of auctions and clickthrough rate. Finally, Figures~\ref{fig:minmax_bctr_sctr_heatmaps}and\ref{fig:minmax_bctr_sctr_heatmaps30} showcase the performance of the baseline bidder on quantiles with fixed clickthrough rates and fixed conversion rates. Cells in the heatmaps which are non-positive in the minimum cases and positive in the maximum cases might be considered ``borderline sparse'', as they pose varying levels of difficulty for the baseline model rather than consistent difficulty or ease to achieve profit.


\section{Experiments with RL Training}\label{app:RL_training}

We employ RLlib \cite{liang2018rllib} as a wrapper for our \env{} to facilitate experimentation and streamline evaluating different RL algorithms environments. RLlib is a popular library that provides an extensive collection of tools and algorithms for RL research. The modular and scalable design of RLlib enables rapid prototyping, customization, and parallelization across multiple CPUs and GPUs, which is crucial for handling computationally demanding tasks in Deep RL research. By integrating our environment with RLlib, we not only ease the adoption of \env{} for the RL community but also enable seamless comparison and integration of various RL algorithms, thereby accelerating research progress in this domain. 

\textbf{Campaign size-} In our experiments, we have specifically chosen a campaign size of 100 keywords. While this campaign size may differ from typical real-world scenarios, it has been selected for focused experimentation and analysis. It is important to note that our choice of a smaller campaign size of 100 keywords is driven by research considerations such as computational constraints, the desired level of complexity for the training process, and the specific objectives we aim to achieve within the scope of our experiments. We recognize that campaign sizes in actual industry practices can vary significantly, spanning from hundreds to thousands or more keywords per campaign. Our decision to utilize a campaign size of 100 keywords serves the purpose of targeted investigation and facilitates in-depth analysis within our research framework.  
\section{Hyperparameters and Network Architectures} \label{app:HP}

For each Deep RL algorithm, we used the recommended hyperparameter settings from the respective original publications and, when necessary, fine-tuned them to achieve optimal performance within the \env{} environment. To ensure a fair comparison, we maintained a consistent set of hyperparameters for each algorithm across all experiments.

\begin{table}[h!]
\centering
\begin{tabular}{|l|c|} 
\hline
\textbf{Parameter} & \textbf{Value} \\
\hline
Actor learning rate & $1e-3$\\ 
\hline
Critic learning rate & $1e-3$\\ 
\hline
Optimizer & Adam\\ 
\hline
Train batch size & $2048$\\ 
\hline
Discount factor (gamma) & $0.95$\\ 
\hline
Soft update factor (tau) & $5e-3$\\ 
\hline
Policy noise & $0.2$\\ 
\hline
Noise clip & $0.5$\\ 
\hline
Policy update frequency & $500$\\ 
\hline
Actor and Critic hidden layers & $400 \times 300$\\ 
\hline
Activation function (actor and critic) & ReLU\\ 
\hline
Replay buffer capacity & $1e-6$\\ 
\hline
Initial sample size before learning & $10000$\\ 
\hline
Exploration configuration & $\{ \text{type: Gaussian Noise}, \text{stddev}: 0.1\}$\\ 
\hline
\end{tabular}
\caption{Hyperparameters and Network Structure of TD3 Algorithm}
\label{table:TD3_hyperparameters}
\end{table}

\begin{table}[h!]
\centering
\begin{tabular}{|l|c|} 
\hline
\textbf{Parameter} & \textbf{Value} \\
\hline
Learning rate & $1e-4$\\ 
\hline
Optimizer & SGD\\ 
\hline
SGD mini-batch size & $64$\\ 
\hline
Number of SGD iterations & $20$\\ 
\hline
Train batch size & $2048$\\ 
\hline
Discount factor (gamma) & $0.995$\\ 
\hline
Lambda & $0.95$\\ 
\hline
PPO clip parameter & $0.5$\\ 
\hline
Hidden layers & $32 \times 32$\\ 
\hline
Activation function & ReLU\\ 
\hline
KL coefficient & $1.0$\\ 
\hline
\end{tabular}
\caption{Hyperparameters and Network Structure of PPO Algorithm}
\label{table:PPO_hyperparameters}
\end{table}

\begin{table}[h!]
\centering
\begin{tabular}{|l|c|} 
\hline
\textbf{Parameter} & \textbf{Value} \\
\hline
Actor Learning rate & $1e-3$\\ 
\hline
Critic Learning rate & $1e-3$\\ 
\hline
Optimizer & Adam\\ 
\hline
Train batch size & $2048$\\ 
\hline
Microbatch batch size & $32$\\ 
\hline
Discount factor (gamma) & $0.99$\\ 
\hline
Lambda & $0.99$\\ 
\hline
Gradient clip parameter & $1.0$\\ 
\hline
Entropy coefficient & $0.01$\\ 
\hline
Value function loss coefficient & $0.5$\\ 
\hline
Actor and Critic hidden layers & $256 \times 256$\\ 
\hline
Activation function & ReLU\\ 
\hline
\end{tabular}
\caption{Hyperparameters and Network Structure of A2C Algorithm}
\label{table:A2C_hyperparameters}
\end{table}

\section{Societal Impact and Limitations} \label{app:soc-lim}
Machine learning models operating in production environments are known to potentially amplify inherent algorithmic bias, a fact extensively documented in literature \cite{milano2021epistemic,birhane2021algorithmic}. In developing \env{}, we have strived to minimize the incorporation of any such biases. However, it is imperative to remain vigilant about the potential unintended consequences these algorithms might engender when operating within the \env{} environment. Emphasis must be placed on rigorous evaluation methodologies and mechanisms that actively mitigate the risk of harm to end users, recognizing the profound societal implications that arise from biased algorithmic decisions.

\section{Compute} \label{app:compute}

Baseline experiments were conducted on Amazon Web Services (AWS) EC2 \texttt{r5.large} instances, which provide the computational resources listed in Table \ref{table:baseline-compute}.
\begin{table}[h]
\centering
\caption{Baseline Model Experiment Compute Specifications}
\label{table:baseline-compute}
\begin{tabular}{|l|l|}
\hline
\textbf{Parameter} & \textbf{Value} \\
\hline
Number of vCPUs & 2 \\
Memory (GiB) & 16.0 \\
Memory per vCPU (GiB) & 8.0 \\
Physical Processor & Intel Xeon Platinum 8175 \\
Clock Speed (GHz) & 3.1 \\
CPU Architecture & x$86\_64$ \\
Number of GPUs & 0 \\
GPU Architecture & none \\
GPU Compute Capability & 0 \\
\hline
\end{tabular}
\end{table}

Deep RL model training experiments were conducted on Amazon Web Services (AWS) EC2 \texttt{m5.12xlarge} instances, which provide the computational resources listed in Table \ref{table:rl-compute}.
\begin{table}[h]
\centering
\caption{Deep RL Models Training Compute Specifications}
\label{table:rl-compute}
\begin{tabular}{|l|l|}
\hline
\textbf{Parameter} & \textbf{Value} \\
\hline
Number of vCPUs & 48 \\
Memory (GiB) & 192.0 \\
Memory per vCPU (GiB) & 4.0 \\
Physical Processor & Intel Xeon Platinum 8175 \\
Clock Speed (GHz) & 3.1 \\
CPU Architecture & x$86\_64$ \\
Number of GPUs & 0 \\
GPU Architecture & none \\
GPU Compute Capability & 0 \\
\hline
\end{tabular}
\end{table}

Table \ref{table:experiment-runtime} presents an overview of the approximate compute resources utilized for all experiments. The table includes the runtime for each experiment, which represents the estimated time required for completion. To calculate the runtime, the evaluation for a single instance of the baseline model was performed using the inline \texttt{\%timeit -r7 -n10} in a Jupyter notebook. The evaluation involved initializing quantiles for the environment, loading them, running the baseline model, and performing necessary computations to estimate the optimal profit used for NCP and AKNCP calculations. Time increases proportionally to the number of keywords sampled, while increasing the number of auctions for the same number of keywords adds a comparatively much smaller computational burden due to the vectorized computation of each keyword's auction outcomes.

\begin{table}[h]
\centering
\caption{Experiment Runtime}
\label{table:experiment-runtime}
\begin{tabular}{|p{8cm}|l|}
\hline
\textbf{Experiment} & \textbf{Runtime} \\
\hline
Baseline Sparsity Heatmap ($16$ seeds/environment on $100$ environment set-ups w/ $100$ keywords each) & $\sim$ 1040 minutes \\
Baseline (Single Stationary Dense instance w/ $100$ keywords for 60 days) & $\sim$ $42.5 \pm 2.07$ seconds \\
Baseline (Single Stationary Denser instance with $m=1024,\,p=0.8$ w/ $100$ keywords for 60 days) & $\sim$ $105$ seconds\\
Baseline (Single Stationary Dense instance w/ $1000$ keywords for 60 days) & $\sim$ $424$ seconds \\
Baseline (Single Stationary Sparse instance w/ $100$ keywords for 60 days) & $\sim$ $27.9 \pm 1.49$ seconds \\
Baseline (Single Non-stationary Dense instance w/ $100$ keywords for 60 days) & $\sim$ $38.2 \pm 0.278$ seconds \\
Baseline (Single Non-Stationary Sparse instance w/ $100$ keywords for 60 days) & $\sim$ $25.1 \pm 0.0863$ seconds\\
Deep RL Stationary Dense & $\sim$ 380 minutes \\
Deep RL Stationary Sparse & $\sim$ 520 minutes \\
Deep RL Non-Stationary Dense & $\sim$ 744 minutes \\
\hline
\end{tabular}
\end{table}

\end{document}